\documentclass{article}

\usepackage{microtype}
\usepackage{graphicx}
\usepackage{subcaption}
\usepackage{booktabs} % for professional tables

\usepackage{hyperref}

\usepackage[preprint]{icml2026}

\usepackage{amsmath}
\usepackage{amssymb}
\usepackage{mathtools}
\usepackage{amsthm}

\usepackage[capitalize,noabbrev]{cleveref}

\usepackage{float}
\usepackage{cancel}

\usepackage{algorithm}
\usepackage{algorithmic}
\usepackage{amsmath}

\usepackage[algo2e]{algorithm2e}
\usepackage{hyperref}

\usepackage{tikz}
\usepackage{amsmath}
\usetikzlibrary{positioning, shapes.geometric, arrows.meta}

\theoremstyle{plain}

\theoremstyle{definition}

\theoremstyle{remark}

\usepackage[textsize=tiny]{todonotes}

\usepackage{enumerate}

\usepackage{enumitem}

\newcommand{\ie}{i.e.\@\xspace}

\newcommand{\mA}{\boldsymbol{A}}

\newcommand{\mD}{\boldsymbol{D}}

\newcommand{\mE}{\boldsymbol{E}}

\newcommand{\mF}{\boldsymbol{F}}

\newcommand{\mG}{\boldsymbol{G}}

\newcommand{\mH}{\boldsymbol{H}}

\newcommand{\mI}{\boldsymbol{I}}

\newcommand{\mM}{\boldsymbol{M}}

\newcommand{\mO}{\boldsymbol{O}}

\newcommand{\mQ}{\boldsymbol{Q}}

\newcommand{\mR}{\boldsymbol{R}}

\newcommand{\mS}{\boldsymbol{S}}

\newcommand{\mT}{\boldsymbol{T}}

\newcommand{\mU}{\boldsymbol{U}}
\newcommand{\vu}{\boldsymbol{u}}
\newcommand{\mV}{\boldsymbol{V}}
\newcommand{\vv}{\boldsymbol{v}}
\newcommand{\mW}{\boldsymbol{W}}

\newcommand{\mX}{\boldsymbol{X}}

\newcommand{\mY}{\boldsymbol{Y}}

\newcommand{\mZ}{\boldsymbol{Z}}

\newcommand{\fP}{\boldsymbol{\mathcal{P}}}

\newcommand{\mSigma}{\boldsymbol{\Sigma}}

\icmltitlerunning{Delving into Muon and Beyond: Deep Analysis and Extensions}

\begin{document}

\twocolumn[
  \icmltitle{Delving into Muon and Beyond: Deep Analysis and Extensions}
  % It is OKAY to include author information, even for blind submissions: the
  % style file will automatically remove it for you unless you've provided
  % the [accepted] option to the icml2026 package.

  % List of affiliations: The first argument should be a (short) identifier you
  % will use later to specify author affiliations Academic affiliations
  % should list Department, University, City, Region, Country Industry
  % affiliations should list Company, City, Region, Country

  % You can specify symbols, otherwise they are numbered in order. Ideally, you
  % should not use this facility. Affiliations will be numbered in order of
  % appearance and this is the preferred way.
  \icmlsetsymbol{equal}{*}
  \icmlsetsymbol{comp}{$\dagger$}

  \begin{icmlauthorlist}
    \icmlauthor{Xianbiao Qi}{xxx,equal}
    \icmlauthor{Marco Chen}{yyy,equal}
    \icmlauthor{Jiaquan Ye}{xxx}
    \icmlauthor{Yelin He}{xxx}
    \icmlauthor{Rong Xiao}{xxx}

  \end{icmlauthorlist}

  %\icmlaffiliation{yyy}{Tsinghua University, Beijing, China}
  %\icmlaffiliation{xxx}{Intellifusion Inc., Shenzhen, China}
  \icmlaffiliation{yyy}{Tsinghua University}
  \icmlaffiliation{xxx}{Intellifusion Inc.}

  \icmlcorrespondingauthor{Xianbiao Qi}{qixianbiao@gmail.com}

  % You may provide any keywords that you find helpful for describing your
  % paper; these are used to populate the "keywords" metadata in the PDF but
  % will not be shown in the document
  \icmlkeywords{Machine Learning, ICML}

  \vskip 0.3in
]

% this must go after the closing bracket ] following \twocolumn[ ...

% This command actually creates the footnote in the first column listing the
% affiliations and the copyright notice. The command takes one argument, which
% is text to display at the start of the footnote. The \icmlEqualContribution
% command is standard text for equal contribution. Remove it (just {}) if you
% do not need this facility.

% Use ONE of the following lines. DO NOT remove the command.
% If you have no special notice, KEEP empty braces:
%\printAffiliationsAndNotice{}  % no special notice (required even if empty)
% Or, if applicable, use the standard equal contribution text:
\printAffiliationsAndNotice{\icmlEqualContribution}

\begin{abstract}
The Muon optimizer has recently attracted considerable attention for its strong empirical performance and use of orthogonalized updates on matrix-shaped parameters, yet its underlying mechanisms and relationship to adaptive optimizers such as Adam remain insufficiently understood.
In this work, we aim to address these questions through a unified spectral perspective. Specifically, we view Muon as the \( p = 0 \) endpoint of a family of spectral transformations of the form \( \boldsymbol{U} \boldsymbol{\Sigma}^{p} \boldsymbol{V}^{\top} \), and consider additional variants with \( p = \frac{1}{2} \), \( p = \frac{1}{4} \), and \( p = 1 \). These transformations are applied to both first-moment updates, as in momentum SGD, and to root-mean-square (RMS) normalized gradient updates as in Adam. To enable efficient computation, we develop a coupled Newton iteration that avoids explicit singular value decomposition. Across controlled experiments, we find that RMS-normalized updates yield more stable optimization than first-moment updates. Moreover, while spectral compression provides strong stabilization benefits under first-moment updates, the Muon update (\( p = 0 \)) does not consistently outperform Adam. These results suggest that Muon is best understood as an effective form of spectral normalization, but not a universally superior optimization method. Our source code will be released at \url{https://github.com/Ocram7/BeyondMuon}.
\end{abstract}

\section{Introduction}
\label{sec:intro}
Optimization methods~\cite{sgd_robbins1951stochastic, msgd_nesterov1983method, adagrad_duchi2011adaptive, rmsprop_hinton2012neural, adam_adam2014method, adamw_loshchilov2019decoupled, shampoo_gupta2018shampoo, jordan_jordan2024muon} are a cornerstone of modern deep learning.
Among them, Adam~\cite{adam_adam2014method} and its variants~\cite{adamw_loshchilov2019decoupled, adafactor_shazeer2018adafactor} are widely adopted due to their strong empirical stability and robustness across architectures and scales.
By combining momentum with variance-based normalization, Adam incorporates second-moment information while remaining computationally efficient, and has consequently become the default optimizer in large-scale neural network training~\cite{llama_touvron2023llamaopenefficientfoundation, llama2_touvron2023llama2openfoundation, llama3_grattafiori2024llama3herdmodels, deepseek_deepseekai2025deepseekr1incentivizingreasoningcapability, qwen_yang2025qwen3technicalreport}.

Recently, Muon~\cite{jordan_jordan2024muon} has attracted growing attention as an alternative optimization strategy, particularly in large language models~\cite{kimi_kimiteam2025kimik2openagentic}.
Unlike most conventional optimizers that operate element-wise, Muon applies matrix-level transformations to gradients, enforcing orthogonality through a spectral operation.
Empirical evidence suggests that Muon can improve convergence behavior, leading to growing interest in the community. Despite growing adoption, the effectiveness and role of Muon remain insufficiently understood.
Existing work has either focused on empirical performance gains~\cite{jordan_jordan2024muon, kimi_kimiteam2025kimik2openagentic} or on mathematical analysis~\cite{newhouse_newhouse2025understanding, bernstein_bernstein2025deriving, sam_buchanan2025mmuonadmm}, but these studies are often accompanied by additional techniques such as QK-Norm or QK-Clip~\cite{kimi_kimiteam2025kimik2openagentic}.
Consequently, it remains unclear to what extent the reported improvements can be attributed to Muon itself, how Muon relates to established adaptive optimizers such as Adam, and whether Muon can be systematically improved. These ambiguities motivate a careful and controlled re-examination of Muon.

In this work, our goal is to study the effectiveness of the Muon optimizer through a unified spectral framework and controlled empirical evaluation.
We view Muon as the \( p = 0 \) endpoint of a family of spectral transformations of the form
\( \boldsymbol{U} \boldsymbol{\Sigma}^{p} \boldsymbol{V}^{\top} \),
and consider fractional variants with \( p=\tfrac{1}{2} \) and \( p=\tfrac{1}{4} \), as well as \( p=1 \), which recovers a standard gradient update.
We evaluate these transformations under both first-moment updates, as in momentum SGD, and root-mean-square-normalized updates, as in Adam.
To make fractional spectral updates practical, we introduce a coupled Newton-Schulz iteration method that computes
\( \mU \mSigma^{\frac{1}{2}} \mV^{\top} \) and
\( \mU \mSigma^{\frac{1}{4}} \mV^{\top} \)
using only matrix multiplications, avoiding explicit SVD.

To ensure fair and interpretable comparisons, we design our experiments to be highly controlled.
We explicitly decouple matrix and vector learning rates for all methods, disable weight decay, and avoid auxiliary techniques such as QK-Norm and QK-Clip.

Under these settings, our experiments yield three key findings: (i) Muon (\(p=0\)) is crucial for stabilizing first-moment updates and substantially improves robustness over mSGD; (ii) when applied to second-moment-normalized updates, Muon-like orthogonalization does not outperform Adam and appears inferior to partial spectral compression; and (iii) across spectral variants, Adam-style second-moment methods are consistently stronger than first-moment ones.

Our main contributions are summarized as follows:
\begin{itemize}[leftmargin=*]
\item We introduce a unified spectral framework \( \mU \mSigma^{p} \mV^{\top} \) that places Muon as the \( p=0 \) endpoint and propose intermediate variants with \( p=\tfrac{1}{2} \) and \( p=\tfrac{1}{4} \).
\item We develop a coupled Newton-Schulz iteration method that enables efficient computation of the fractional spectral updates \(p=\tfrac{1}{2} \) and \( p=\tfrac{1}{4} \) without explicit SVD.
\item We provide a rigorous and controlled empirical comparison between Muon, Adam, and their spectral variants across first and second-moment-normalized updates, isolating spectral effects from other confounding techniques.
\end{itemize}

\section{Related Work}
\textbf{Optimizers before Transformer.}
Before the Transformer~\cite{transformer_vaswani2017attention}, CNNs~\cite{cnn_lecun2002gradient, resnet_he2016deep} and shallow recurrent models such as LSTMs~\cite{lstm_hochreiter1997long} dominated neural network design. These architectures were typically of moderate depth with limited cumulative nonlinearity, resulting in relatively small variation in layer-wise Lipschitz constants~\cite{lipsformer_qi2023lipsformer, qi_qi2023understanding}. Consequently, training with a global learning rate was generally effective.
Under this regime, classical stochastic optimization methods~\cite{book_leon_bottou2018optimization} were sufficient. Stochastic Gradient Descent (SGD)~\cite{sgd_robbins1951stochastic} and its variants~\cite{msgd_nesterov1983method, svrg_johnson2013accelerating}, often combined with simple learning rate schedules, were the standard choices.

Although adaptive methods such as Adagrad~\cite{adagrad_duchi2011adaptive}, RMSProp~\cite{rmsprop_hinton2012neural}, and Adam~\cite{adam_adam2014method} had been proposed, they were not widely adopted at the time. This can be attributed to two main factors. First, weight decay was primarily viewed as a regularization technique, and its interaction with adaptive optimizers was poorly understood; in particular, coupling weight decay with gradients rather than weights significantly degraded the performance of Adam-type methods~\cite{adamw_loshchilov2019decoupled}. Second, in CNN-based models, inter-layer differences in Lipschitz constants were insufficient to necessitate per-parameter adaptivity, and momentum SGD already achieved strong empirical performance. As a result, Adam did not consistently outperform mSGD in this setting.

\textbf{Optimizers after Transformer.}
The introduction of Transformers~\cite{transformer_vaswani2017attention} fundamentally altered the optimization landscape of deep learning. Compared to earlier architectures, Transformer-based models are substantially deeper and exhibit stronger nonlinearities~\cite{swishglu_shazeer2020glu}, arising from attention mechanisms~\cite{transformer_vaswani2017attention}, residual connections~\cite{resnet_he2016deep}, and normalization layers~\cite{layernorm_ba2016layer}. Consequently, different layers and parameter groups often exhibit markedly different curvature and gradient statistics, often leading to training instability and suboptimal convergence. 

AdamW~\cite{adamw_loshchilov2019decoupled} resolved the improper coupling between weight decay and adaptive learning rates in Adam, leading to significantly improved performance on Transformer models. As a result, AdamW consistently outperforms momentum SGD in large-scale Transformer training and has become a standard optimizer choice. Subsequent work has proposed a range of alternative optimizers, including Adafactor~\cite{adafactor_shazeer2018adafactor}, SignSGD~\cite{signsgd_bernstein2018signsgd}, LAMB~\cite{you_you2019large}, Adan~\cite{adan_xie2024adanadaptivenesterovmomentum}, Lion~\cite{lion_chen2023symbolic}, Sophia~\cite{sophia_liu2024sophiascalablestochasticsecondorder}, Mars~\cite{mars_yuan2025mars}, and related variants~\cite{adam_mini_zhang2025adamminiusefewerlearning, cautious_liang2025cautious}. These methods primarily target improved stability, efficiency, or convergence in large-scale, highly nonlinear training regimes.

\textbf{Matrix-based Optimizers.}
Matrix-based optimizers exploit the structured geometry of parameters by operating directly on matrix-valued variables, rather than treating parameters as independent scalars. Early work in this direction focused on making second-order optimization tractable via structured approximations. K-FAC~\cite{kfac_martens2015optimizing} introduced a Kronecker-factored approximation of the Fisher information matrix, enabling efficient layer-wise preconditioning, while Shampoo~\cite{shampoo_gupta2018shampoo} extended this idea by applying Kronecker factorizations along multiple matrix dimensions, leading to improved stability. These methods share the goal of incorporating curvature information through structured matrix approximations.

More recently, Muon~\cite{jordan_jordan2024muon} has emerged as a matrix-based optimizer that departs from explicit curvature modeling. Muon directly updates matrix-valued parameters via orthogonalization and spectral transformations computed using Newton--Schulz iterations~\cite{newton_schulz_schulz1933iterative, higham_higham2008functions}, relying solely on matrix multiplications. Since its introduction, Muon has inspired a growing body of follow-up work~\cite{kimi_kimiteam2025kimik2openagentic, adamuon_si2025adamuonadaptivemuonoptimizer, what_frans2025really, suweijie_su2025isotropic, polargrad_lau2025polargrad}, including analyses that clarify its numerical linear algebra foundations~\cite{bernstein_bernstein2025deriving, newhouse_newhouse2025understanding} and geometric interpretations that view it as implicit manifold-aware optimization~\cite{bernstein_bernstein2025manifolds, sam_buchanan2025mmuonadmm}.

\textbf{Remarks.}
Unlike several previous works on optimizer benchmarking~\cite{wen_optimizer_wen2025fantastic, what_frans2025really}, our study is characterized by the following features: (1) we introduce a unified spectral-based formulation of the form $\mU \mSigma^{p} \mV^{\top}$ with $p \in [0,1]$, and instantiate it with four concrete choices, $p\in\{1,\tfrac12,\tfrac14,0\}$. For $\mU \mSigma^{\frac{1}{2}} \mV^{\top}$ and $\mU \mSigma^{\frac{1}{4}} \mV^{\top}$, we develop a coupled Newton--Schulz algorithm to solve them. (2) we experimently show that matrix-based optimizers with second-moment information are stronger than their pure first-moment counterparts. (3) we design our experiments to be highly controlled.
We explicitly decouple matrix and vector learning rates for all methods, disable weight decay, and avoid auxiliary techniques. This allows us to isolate the intrinsic capabilities of each optimizer, enabling a fair and transparent comparison.

\section{Delving into Muon and Beyond}
\label{sec:algorithm}
\subsection{Baselines: Adam and Muon}

As in Muon, our study focuses exclusively on \emph{matrix-shaped parameters}
\(\mW \in \mathbb{R}^{m \times n}\); all vector parameters are optimized using standard Adam.
In this subsection, we briefly review Adam and Muon and fix notation.

\paragraph{Adam.}Adam~\cite{adam_adam2014method} is a popular stochastic optimizer that combines momentum with adaptive, second-moment-based normalization.
Given the gradient matrix
\(\mG_t = \nabla_{\mW_t}\ell(\mW_t)\) at iteration $t$,
Adam maintains exponentially decaying estimates of the first and second moments:
\begin{align}
\mM_t &= \beta_1 \mM_{t-1} + (1 - \beta_1)\mG_t, \label{eq:adam-m} \\
\mV_t &= \beta_2 \mV_{t-1} + (1 - \beta_2)\mG_t^{\,2}, \label{eq:adam-v}
\end{align}
and updates parameters via
\begin{equation}
\mW_{t+1} = \mW_t - \eta_t\, \mM_t \oslash \sqrt{\mV_t}.
\label{eq:adam-update}
\end{equation}
Here, \(\beta_1\) and \(\beta_2\) are the first- and second-moment decay rates, \(\eta_t\) is the learning rate, and \(\oslash\) denotes elementwise division.
From a preconditioning perspective, Adam applies a diagonal matrix \(\mV_t^{-1/2}\) to the momentum \(\mM_t\), normalizing each coordinate by an estimate of its second-moment and yielding bounded, scale-adaptive updates. For brevity, throughout the rest of this work we refer to
$\mM_t \oslash \sqrt{\mV_t}$ as the \emph{RMS-normalized} update, since $\mV_t$
tracks an exponential moving average of squared gradients.

\paragraph{Muon.}Muon~\cite{jordan_jordan2024muon} replaces Adam’s elementwise normalization with a
\emph{spectral normalization} of the update matrix.
Given the momentum matrix \(\mM_t\) with singular value decomposition
\(\mM_t = \mU \mSigma \mV^{\top}\),
Muon defines the update direction as the polar factor
\[
\fP(\mM_t) := \mU \mV^{\top},
\]
which discards the singular values and retains only the left and right singular vectors. Computing the polar factor is typically referred to as matrix orthogonalization.

From an optimization perspective, orthogonalization enforces \emph{unit magnitude along every singular direction}.
In contrast to Adam, which normalizes updates on an elementwise basis via a diagonal preconditioner,
Muon equalizes the strength of updates across all directions of the gradient matrix.
As a result, Muon removes anisotropic scaling in the spectrum of \(\mM_t\), yielding directionally balanced but magnitude-agnostic updates.

\paragraph{Newton--Schulz approximation.}
Computing the polar factor
\(
\fP(\mM_t) = \mU \mV^{\top} = \mM_t(\mM_t^{\top}\mM_t)^{-1/2}
\)
via an explicit SVD is prohibitively expensive for large models.
However, we can approximate the inverse square root using \emph{Newton--Schulz iteration},
which relies only on matrix multiplications. Let
$\mA = \alpha \mM_t^{\top}\mM_t,$
where the scaling constant \(\alpha\) is chosen such that
\(\|\mI - \mA\|_2 \leq 1\).
The iteration
\begin{equation}
\mZ_{k+1}
=
\tfrac{1}{2}\,\mZ_k\bigl(3\mI - \mA \mZ_k^2\bigr), \quad \mZ_0 = \mI,
\label{eq:ns_muon}
\end{equation}
converges quadratically to \( \frac{1}{\sqrt{\alpha}}(\mM_t^{\top}\mM_t)^{-1/2}\).
After \(K\) iterations, the polar factor is approximated as
\begin{equation}
\fP(\mM_t) \approx \sqrt{\alpha} \mM_t \mZ_K.
\end{equation}

In practice, Muon~\cite{jordan_jordan2024muon} employs \autoref{alg:ns5_muon} in \autoref{appendix:NS_by_jordan}, which is a slightly more direct and aggressive version of the Newton-Schulz iteration shown in \autoref{eq:ns_muon}. The official implementation directly computes the polar factor with a procedure that converges for initializations where the singular values are adequately normalized.

\subsection{Momentum $\;\text{vs.}\;$ RMS-Normalized Updates}
\paragraph{Where to apply spectral transforms.}
Muon~\cite{jordan_jordan2024muon} applies its spectral operation to the \emph{first-moment}
momentum $\mM_t$. However, in large-scale Transformer training, Adam-style optimizers that incorporate second-moment statistics generally
outperform momentum SGD-based optimizers that only leverage first-moment information. This observation motivates us to examine whether spectral transformations yield additional benefits when applied to rms-
normalized updates. Concretely, we study the two inputs
\[
\mO_t^{\text{mom}} := \mM_t
\quad \text{and} \quad
\mO_t^{\text{rms}} := \mM_t \oslash \sqrt{\mV_t},
\]
which correspond to the first-moment update shown in \autoref{eq:adam-m} and the rms-normalized update from \autoref{eq:adam-update}.

\textbf{How $\mO_t^{\text{mom}}$ and $\mO_t^{\text{rms}}$ differ.}
The key distinction between these two regimes lies in how update magnitudes are controlled.
In first-order momentum methods, the update $\mM_t$ aggregates gradients over time without
explicit normalization by their scale.
As a result, the norm of the update can grow with accumulated gradient magnitude, especially
under high variance or anisotropic curvature.
Consequently, the update scale is not bounded by the optimizer.

In contrast, RMS-normalized updates explicitly rescale the momentum using second-moment statistics. The update $\mM_t \oslash \sqrt{\mV_t}$ normalizes each coordinate by an estimate of its raw second moment, yielding an update whose magnitude is provably bounded. For Adam-style methods, this bound, $\frac{1-\beta_1}{\sqrt{1-\beta_2}}$~\cite{qi_qi2023understanding}, depends only on the decay rates and is independent of the raw gradient magnitude.
This intrinsic normalization leads to improved numerical stability and more predictable
optimization dynamics.

\subsection{A Spectral Family of Transformations}
We introduce a family of spectral gradient transformations that enables a unified analysis of Muon-inspired spectral methods. Given SVD, $\mO_t = \mU \mSigma \mV^\top$, we define the general spectral transformation
\begin{equation}
\Psi_p(\mO_t) := \mU \mSigma^p \mV^\top, \qquad p\in[0,1].
\label{eq:spectral_transform}
\end{equation}
The exponent $p$ directly controls how the singular spectrum is rescaled. Writing the SVD in its rank-one form, we obtain
\begin{equation}
    \Psi_p(\mO_t)
    \;=\;
    \sum_{i=1}^{d} \sigma_i^{\,p}\, \vu_i \vv_i^\top,
\label{eq:svd_p}
\end{equation}
which makes it explicit that decreasing $p$ progressively compresses the singular values $\{\sigma_i\}$ while preserving the singular directions $\{\vu_i, \vv_i\}$.

This formulation subsumes several existing methods and enables systematic comparisons across spectral behaviors. In particular, $p=1$ leaves the input unchanged, recovering standard methods like mSGD and Adam, while $p=0$ maps all nonzero singular values to $1$, yielding an orthogonalized update direction. When the input is the first-moment momentum $\mO_t=\mO_t^{\text{mom}}$, $p=0$ corresponds to Muon.

\subsection{Instantiating the Spectral Family}\label{sec:instantiation}

\paragraph{From a spectral operator to concrete optimizers.}
We instantiate our spectral update family by applying the map $\Psi_p(\cdot)$ from
\autoref{eq:spectral_transform} to one of two matrix-valued inputs:
the first-moment momentum $\mO_t^{\text{mom}}$, or the RMS-normalized update
$\mO_t^{\text{rms}}$.
We consider four spectral exponents
\[
p \in \Bigl\{1,\tfrac12,\tfrac14,0\Bigr\},
\]
where $p=1$ leaves the input unchanged and $p=0$ collapses the spectrum to the polar factor.
The intermediate powers $p=\tfrac12$ and $p=\tfrac14$ provide milder alternatives to Muon's
fully flattened spectrum, yielding a controlled interpolation
\[
\mU\mSigma^{1}\mV^{\top}
\;\rightarrow\;
\mU\mSigma^{1/2}\mV^{\top}
\;\rightarrow\;
\mU\mSigma^{1/4}\mV^{\top}
\;\rightarrow\;
\mU\mSigma^{0}\mV^{\top}.
\]
Combining two inputs with four exponents produces the eight optimizer instances evaluated in this work.

\paragraph{Naming convention.}
Each instance is identified by its \emph{input family} and \emph{spectral exponent}.
We write
\[
\textsc{Base}\mathrm{X}
\;\;:\;\;
\Delta \mW_t \propto \Psi_p(\mO_t^{\cdot}),
\]
where $\textsc{Base}\in\{\textbf{mSGD},\textbf{Adam}\}$ selects
$\mO_t^{\text{mom}}$ or $\mO_t^{\text{rms}}$, and the suffix
$\mathrm{X}\in\{\varnothing,\textbf{S},\textbf{Q},\textbf{Z}\}$ encodes the exponent
\[
\begin{array}{ll}
\varnothing \leftrightarrow p=1 \;(\text{identity}), &
\textbf{S} \leftrightarrow p=\tfrac12 \;(\text{square-root}), \\[2pt]
\textbf{Q} \leftrightarrow p=\tfrac14 \;(\text{quarter-power}), &
\textbf{Z} \leftrightarrow p=0 \;(\text{zero-power / polar}).
\end{array}
\]

\paragraph{Momentum-input family ($\mO_t^{\text{mom}}=\mM_t$).}
Under the momentum input, the four instances are
\[
\textbf{mSGD},\;\textbf{mSGDS},\;\textbf{mSGDQ},\;\textbf{mSGDZ},
\]
corresponding to $p\in\{1,\tfrac12,\tfrac14,0\}$ respectively.
The endpoint \textbf{mSGDZ} is equivalent to \textbf{Muon} since $\Psi_0(\mM_t)=\fP(\mM_t)=\mU\mV^\top$.

\paragraph{RMS-normalized-input family ($\mO_t^{\text{rms}}=\mM_t\oslash\sqrt{\mV_t}$).}
Under the RMS-normalized input, the four instances are
\[
\textbf{Adam},\;\textbf{AdamS},\;\textbf{AdamQ},\;\textbf{AdamZ},
\]
again corresponding to $p\in\{1,\tfrac12,\tfrac14,0\}$ respectively.

\paragraph{Spectral anisotropy control.}
The four spectral variants in each input family share the same singular vectors $(\mU,\mV)$ and differ only in how the singular values are rescaled. Thus, the effect of the spectral transformation can be understood purely as reshaping the anisotropy of the singular spectrum. To quantify anisotropy, we use the spectral condition number
\[
\kappa(\mO_t) \;=\; \frac{\sigma_{\max}}{\sigma_{\min}},
\]
which measures the spread of the singular values. When $\kappa(\mO_t)$ is large, the input is highly ill-conditioned and a small number of
dominant singular directions can disproportionately influence the update.

\begin{figure}[t]
\centering
\resizebox{0.95\columnwidth}{!}{
\begin{tikzpicture}[
    box/.style={
        draw,
        rounded corners,
        inner sep=7pt,
        minimum height=10mm,
        minimum width=8.0cm,
        align=center,
        font=\LARGE
    },
    label/.style={font=\Huge},
    arrow/.style={->, thick, line width=2.0pt}
]
\definecolor{powone}{RGB}{0,0,0}
\definecolor{powhalf}{RGB}{33,113,181}
\definecolor{powquarter}{RGB}{35,139,69}
\definecolor{powzero}{RGB}{160,160,160}

\node[box, draw=powone] (A)
{$[\,9 \quad 4 \quad 1 \quad 0.01 \quad 0.0001\,]$};
\node[label, above=3mm of A, text=powone]
{$U \Sigma^{1} V^{\top}$};

\node[box, draw=powhalf, right=2.8cm of A] (B)
{$[\,3 \quad 2 \quad 1 \quad 0.1 \quad 0.01\,]$};
\node[label, above=3mm of B, text=powhalf]
{$U \Sigma^{\tfrac12} V^{\top}$};
\draw[arrow, draw=powhalf] (A) -- (B);

\node[box, draw=powzero, below=1.8cm of A] (C)
{$[\,1 \quad 1 \quad 1 \quad 1 \quad 1\,]$};
\node[label, below=3mm of C, text=powzero]
{$U \Sigma^{0} V^{\top}$};
\draw[arrow, draw=powzero] (A) -- (C);

\node[box, draw=powquarter, below right=1.8cm and 2.8cm of A] (D)
{$[\,\sqrt{3} \quad \sqrt{2} \quad 1 \quad \sqrt{0.1} \quad \sqrt{0.01}\,]$};
\node[label, below=3mm of D, text=powquarter]
{$U \Sigma^{\tfrac14} V^{\top}$};
\draw[arrow, draw=powquarter] (A) -- (D);

\end{tikzpicture}
}
\caption{Effect of spectral exponent $p$ on the singular spectrum (illustrative numbers). Decreasing $p$ compresses the spectrum: large singular values are damped relative to small ones, and $p=0$ maps all nonzero singular values to $1$.}
\label{fig:spectral-powers}
\end{figure}
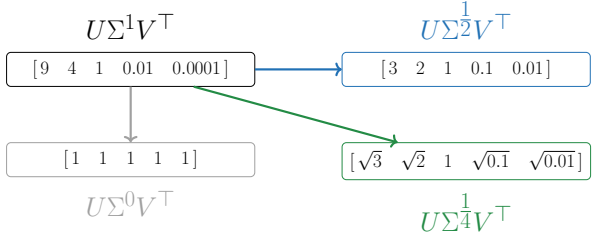

If the input has singular values \(\{\sigma_i\}_{i=1}^{d}\), then \(\Psi_p\) produces singular values \(\{\sigma_i^{p}\}_{i=1}^{d}\),
so the condition number becomes
\[
\kappa_p \;=\; \frac{\sigma_{\max}^{p}}{\sigma_{\min}^{p}} \;=\; \kappa(\mO_t)^{p}.
\]
Therefore, decreasing \(p\) reduces spectral anisotropy:
\[
\kappa_{1} \geq \kappa_{1/2} \geq \kappa_{1/4} \geq \kappa_{0}=1,
\]
Indeed, for the toy spectrum in \autoref{fig:spectral-powers}, $\kappa_{1}=90000$, $\kappa_{1/2}=300$, $\kappa_{1/4}=\sqrt{300}$, and $\kappa_{0}=1$. This viewpoint clarifies how the exponent \(p\) continuously interpolates between the original spectrum (\(p=1\)) and complete spectral flattening (\(p=0\)).

\paragraph{The effect of spectral compression.}As shown in \autoref{fig:spectral-powers}, decreasing \(p\) progressively compresses the \emph{entire} singular-value spectrum. Hence, directions associated with large singular values are attenuated, while those associated with small singular values (less than one) are amplified.

\subsection{Efficient Computation}
\paragraph{Computing $\Psi_{1/2}(\mO_t)$ and $\Psi_{1/4}(\mO_t)$.}A practical challenge in spectral methods is computational efficiency, as direct SVD computation is infeasible for large models. Fortunately, we can rewrite $\Psi_{1/2}(\mO_t)$ in a form that avoids explicit SVD. Let $\mO_t=\mU\mSigma\mV^{\top}$ and assume $m\ge n$. Then
\[
\Psi_{1/2}(\mO_t)=\mU\mSigma^{1/2}\mV^{\top}
\;=\;
\mO_t\,(\mO_t^{\top}\mO_t)^{-1/4}.
\]
Thus, it suffices to compute an inverse fourth root of the symmetric matrix
$\mX := \mO_t^{\top}\mO_t \in \mathbb{R}^{n\times n}.$
We can compute $\mX^{\frac{1}{2}}$ and $\mX^{\frac{-1}{2}}$ efficiently using the coupled Newton-Schulz algorithm discussed below, and obtain $\mX^{-1/4}$ by applying the same procedure to $\mX^{1/2}$ (\ie, computing $(\mX^{1/2})^{-1/2}=\mX^{-1/4}$). Consequently, $\Psi_{1/2}(\mO)$ can be implemented using only matrix multiplications as
$
\Psi_{1/2}(\mO) \;=\; \mO\,\mX^{-1/4}$.
A similar matrix-multiplication-only procedure can be used for the quarter-power spectral transformations.

\paragraph{Coupled Newton-Schulz Iteration.} Coupled Newton-Schulz iteration~\cite{higham_higham2008functions} provides an efficient and numerically stable procedure for simultaneously computing the matrix square root $\mX^{1/2}$ and its inverse $\mX^{-1/2}$.

Starting from the initialization $\mY_0=\mX$ and $\mZ_0=\mI$, the method applies a coupled update that repeatedly refines both quantities using only matrix multiplications.
Specifically, each iteration is defined as
\begin{equation}
    \begin{aligned}
     \mY_{k+1} &= \frac{1}{2}\,\mY_k \bigl(3\mI-\mZ_k\mY_k\bigr), \\
     \mZ_{k+1} &= \frac{1}{2}\,\bigl(3\mI-\mZ_k\mY_k\bigr)\mZ_k, 
    \end{aligned}
\end{equation}
which symmetrically updates $\mY_k$ and $\mZ_k$ through the shared correction term $3\mI-\mZ_k\mY_k$.
When appropriately scaled, this coupled iteration converges quadratically, driving $\mY_k \to \mX^{1/2}$ and $\mZ_k \to \mX^{-1/2}$ simultaneously. 

\begin{algorithm}[t]
\caption{Coupled Newton-Schulz for $\mX^{\frac{1}{2}}$ and $\mX^{\frac{-1}{2}}$}
\label{alg:couple_ns}
\begin{algorithmic}[1]
\REQUIRE $\mX$, number of iterations $K$
\ENSURE $\mX^{\frac{1}{2}}$, $\mX^{-\frac{1}{2}}$

\STATE $\mY_0 \leftarrow \mX$, \quad $\mZ_0 \leftarrow \mI$
\STATE $\alpha \leftarrow \|\mX\|_F$
\STATE $\mY_0 \leftarrow \mY_0 / \alpha$, \quad  $k \leftarrow 0$

\WHILE{$k < K$}
    \STATE $\mT_k \leftarrow 3\mI - \mZ_k \mY_k$
    \STATE $\mY_{k+1} \leftarrow \tfrac{1}{2}\, \mY_k \mT_k$
    \STATE $\mZ_{k+1} \leftarrow \tfrac{1}{2}\, \mT_k \mZ_k$
    \STATE $k \leftarrow k + 1$
\ENDWHILE

\STATE \textbf{return} $\sqrt{\alpha}\,\mY_K,\; \frac{1}{\sqrt{\alpha}}\,\mZ_K$
\end{algorithmic}
\end{algorithm}

In practice, a normalization step based on the Frobenius norm of $\mX$ is applied at initialization to ensure numerical stability, and the final iterates are rescaled accordingly.
In contrast to the standard one-sequence Newton–Schulz iteration, which can be obtained by eliminating one variable from the coupled Newton iteration under a commuting initialization, the coupled formulation simultaneously evaluates $\mY_k$ and $\mZ_k$, making it a more suitable tool for computing matrix roots and inverse roots. \autoref{alg:couple_ns} depicts our method.

\section{Experiments}
\label{sec:experiments}

\subsection{Experimental Settings}\label{sec:experiments:setup}
\paragraph{Setup.}
We conduct our experiments on nanoGPT\footnote{\url{https://github.com/karpathy/nanoGPT}}~\cite{nanogpt_Karpathy2022}, a lightweight GPT-2 training codebase.
We follow the standard GPT-2 configuration: GELU activations and a byte-pair encoding tokenizer with vocabulary size 50{,}257. Our GPT-2 model has 124M parameters.
All runs use sequence length 1024, global batch size 480, and are trained on OpenWebText for 200K optimization steps with warmup for 2{,}000 steps.
We do not use QK-Norm or QK-Clip.
Across runs, we keep all settings fixed and vary only the optimizer and its learning rate.

For Muon, we use the reference implementation\footnote{\url{https://github.com/KellerJordan/Muon}} from~\citet{jordan_jordan2024muon}, which applies the matrix update
$\mW_{t+1} \;=\; \mW_t \;-\; \eta_t \sqrt{\frac{\mathrm{fan\text{-}out}}{\mathrm{fan\text{-}in}}}\,\mU \mV^{\top},$
where $\mU\mV^{\top}$ is the polar factor computed via \autoref{alg:ns5_muon}.

\begin{figure*}[t]
    \centering
    \begin{subfigure}{0.50\textwidth}
        \centering
        \includegraphics[width=\linewidth]{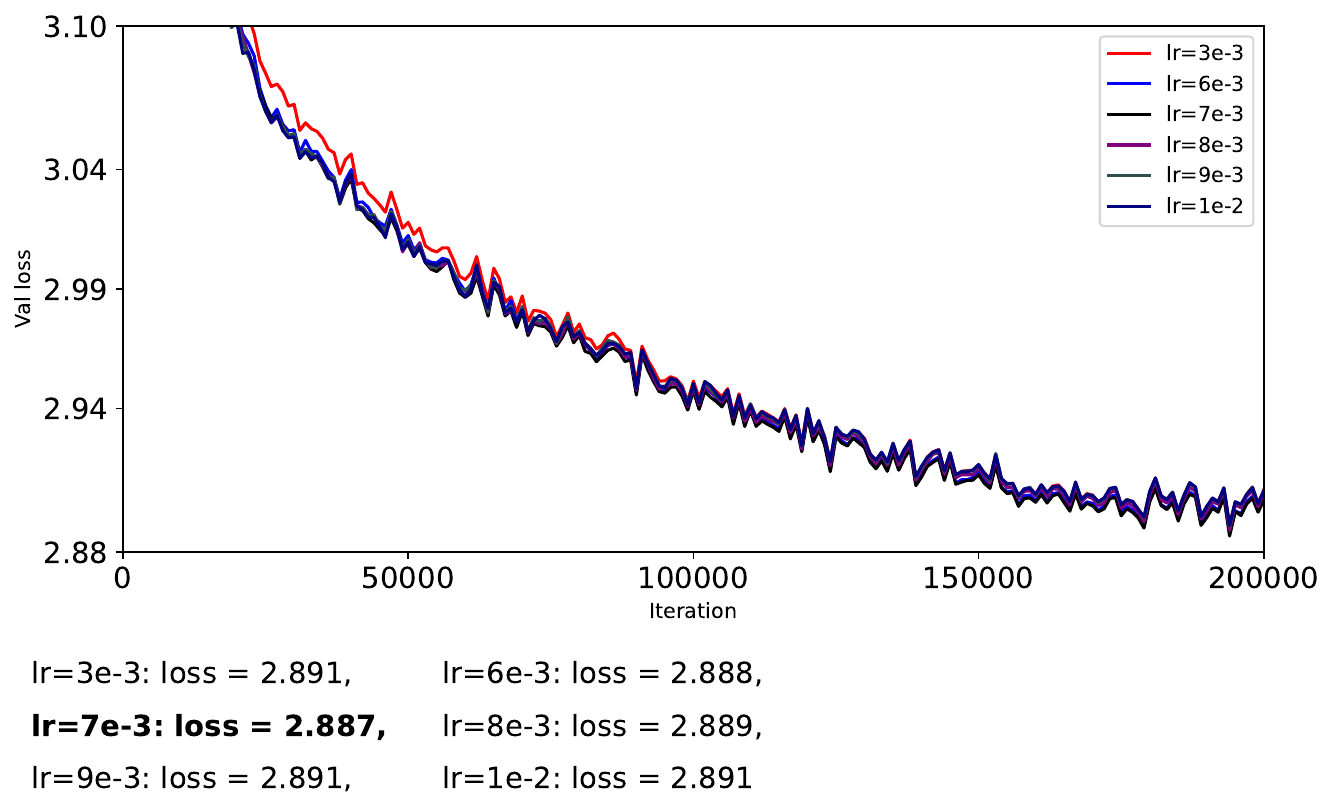}
        \caption{mSGDZ/Muon ($\mU \mSigma^{0}\mV^{\top}$)}
        \label{fig:msgd_p0}
    \end{subfigure}
    \begin{subfigure}{0.49\textwidth}
        \centering
        \includegraphics[width=\linewidth]{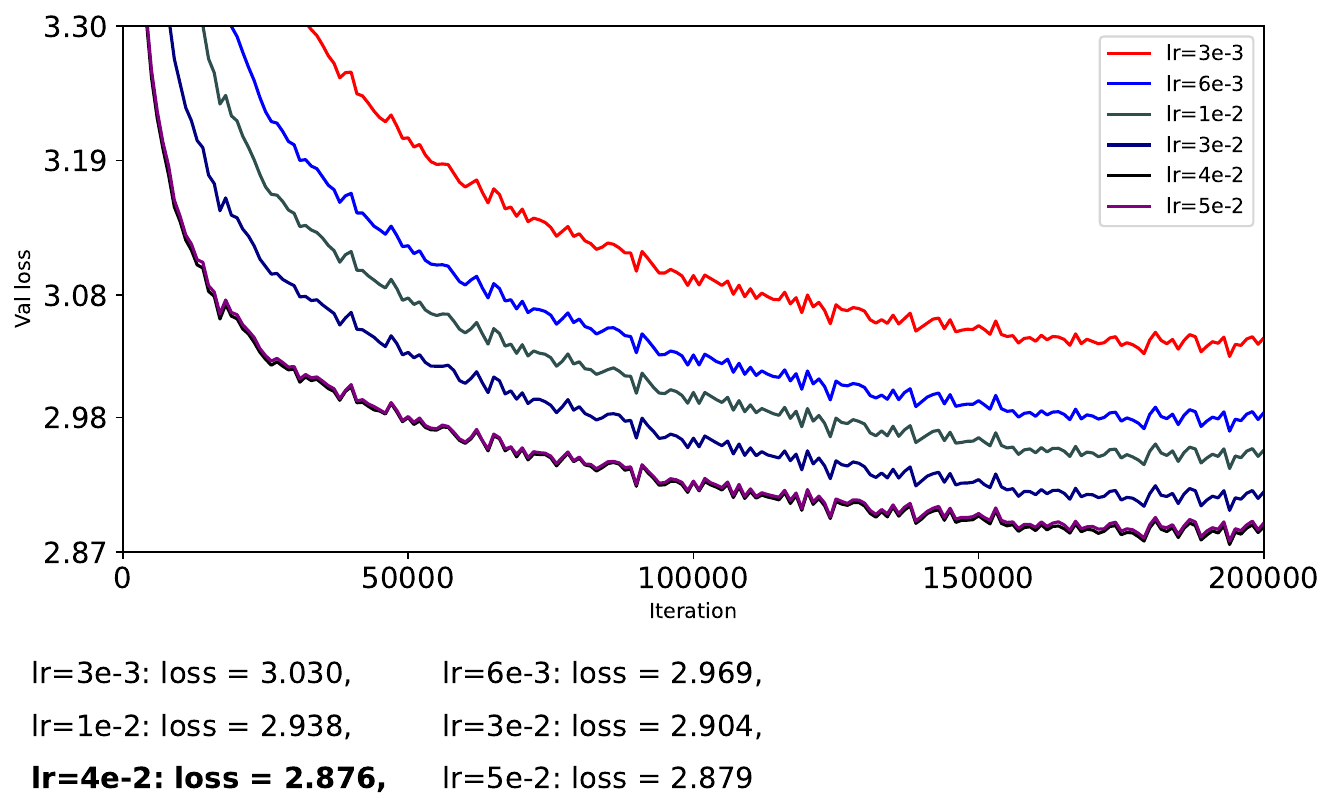}
        \caption{mSGDQ ($\mU \mSigma^{\tfrac{1}{4}}\mV^{\top}$)}
        \label{fig:msgd_p1_4}
    \end{subfigure}
    \begin{subfigure}{0.49\textwidth}
        \centering
        \includegraphics[width=\linewidth]{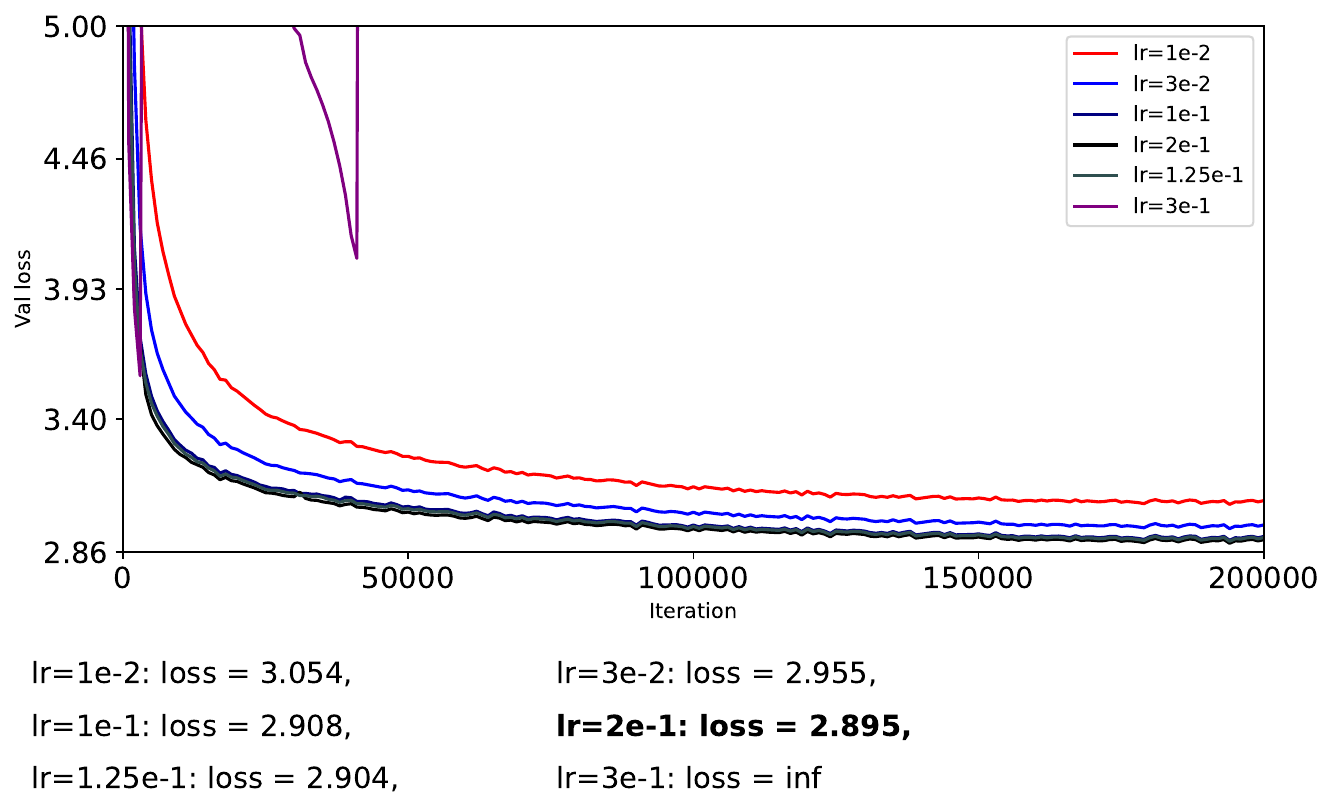}
        \caption{mSGDS ($\mU \mSigma^{\tfrac{1}{2}}\mV^{\top}$)}
        \label{fig:msgd_p1_2}
    \end{subfigure}
    \begin{subfigure}{0.49\textwidth}
        \centering
        \includegraphics[width=\linewidth]{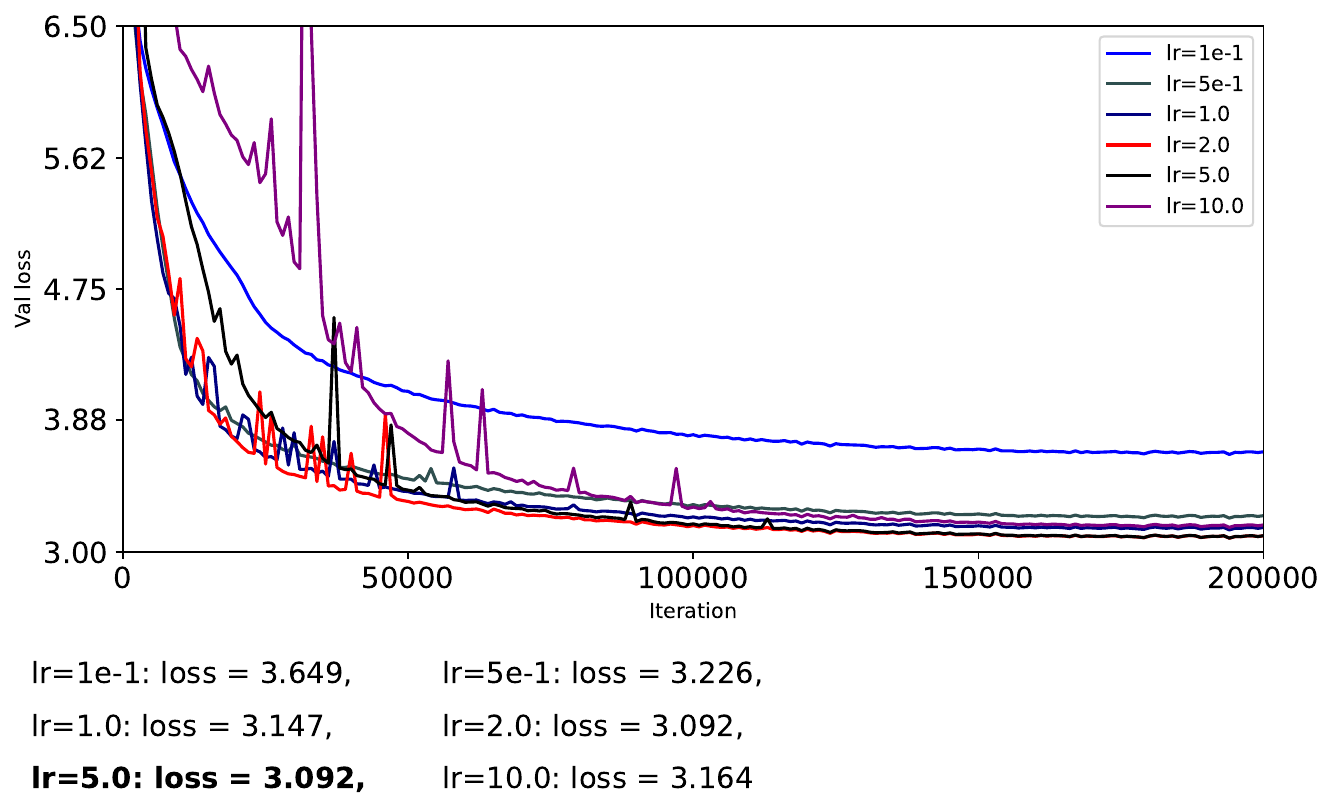}
        \caption{mSGD ($\mU \mSigma^{1}\mV^{\top}$)}
        \label{fig:msgd_p1}
    \end{subfigure}
    \caption{
        Overall comparison across four optimizers (mSGDZ/Muon, mSGDQ, mSGDS and mSGD) based on the \textbf{first-moment momentum $\mM_t$}.
        Each subfigure corresponds to a different optimizer.
    }
    \label{fig:msgd_all}
\end{figure*}
\paragraph{Controlled comparison.}
To ensure fair and interpretable comparisons, we isolate the optimizer's effect on \emph{matrix-shaped} parameters and, crucially, \emph{decouple matrix and vector learning rates for \textbf{all} optimizers}.

Muon is typically evaluated with separate learning rates for matrix and vector parameters, whereas standard Adam baselines often share a single global learning rate. To eliminate this mismatch, we follow Muon and apply Adam to all vector-valued parameters for every method using a fixed learning rate $\text{lr}_{\text{vec}}=3\times 10^{-4}$. For matrix parameters, each optimizer is tuned with its own learning rate $\text{lr}_{\text{mat}}$.

We also disable weight decay for all runs (wd$=0$). In standard implementations, weight decay is coupled to the learning rate through the update
$\mW_{t+1}
=
\mW_t
-\eta_t\,\Delta\mW_t
-\eta_t \lambda \mW_t,
$
so jointly tuning $(\eta_t,\lambda)$ introduces an additional degree of freedom that complicates controlled cross-optimizer comparisons. In addition, we avoid auxiliary stabilization techniques such as QK-Norm and QK-Clip. All remaining hyperparameters are held fixed across methods; in particular, we set $\beta_1=0.9$ and $\beta_2=0.95$ throughout.

Overall, this design decouples the matrix learning rate from other hyperparameters, including vector learning rates and weight decay, so performance differences are more directly attributable to the matrix-parameter update rule itself.

\subsection{Learning Rate Tuning Protocol}
We tune the matrix-parameter learning rate $\text{lr}_{\text{mat}}$ for each optimizer using a two-stage procedure: a coarse logarithmic sweep to identify a stable scale, followed by a local refinement within that scale.

\paragraph{Coarse search.}
We first sweep $\text{lr}_{\text{mat}}$ on a logarithmic grid spanning several orders of magnitude to locate the region where the optimizer transitions from divergence to effective learning. For instance, for AdamW we evaluate
$\{10^{-1},\,10^{-2},\,10^{-3},\,10^{-4},\,10^{-5}\}.$
This stage intends to identify the correct order of magnitude rather than the exact optimum. 
Since most candidates in this sweep diverge, we do not plot the resulting curves.

\begin{figure*}[t]
    \centering
    \begin{subfigure}{0.50\textwidth}
        \centering
        \includegraphics[width=\linewidth]{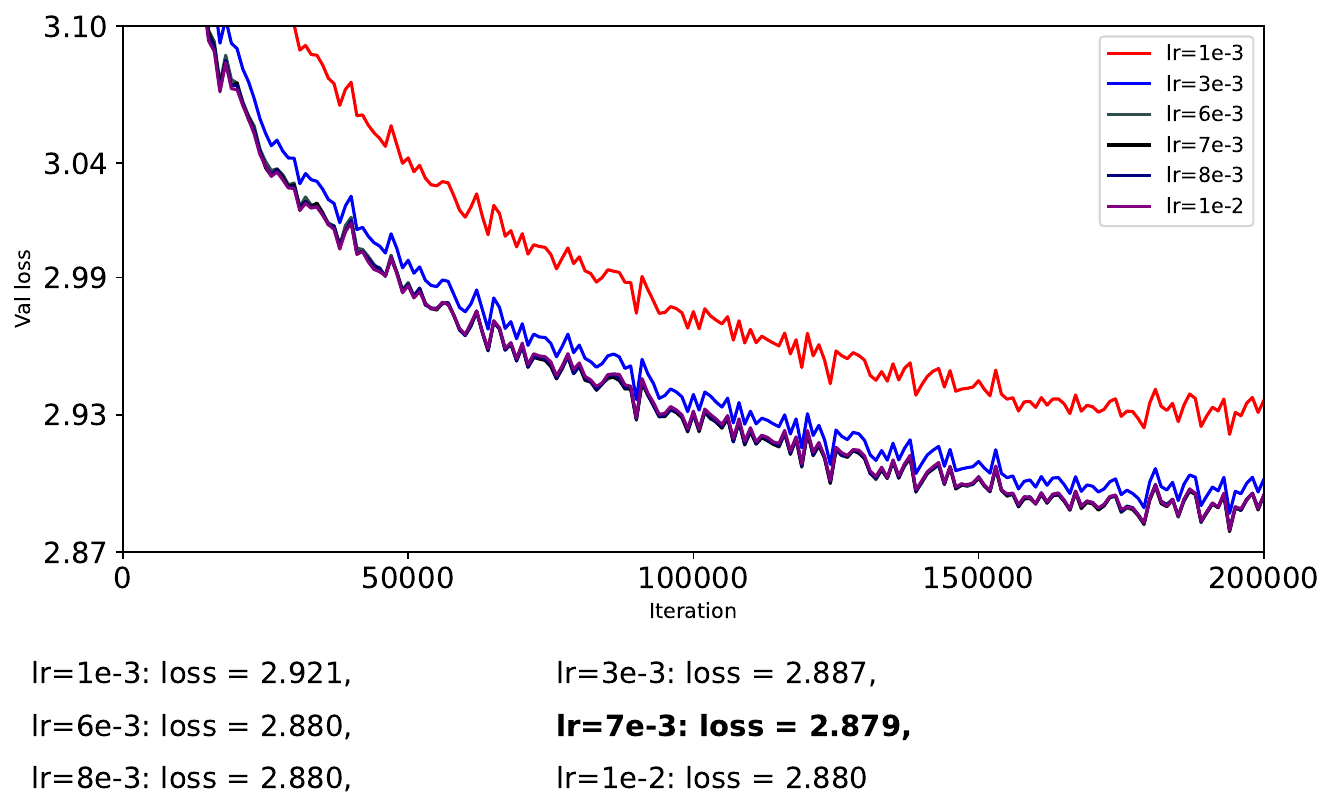}
        \caption{AdamZ ($\mU \mSigma^{0}\mV^{\top}$)}
        \label{fig:adam_p0}
    \end{subfigure}
    \begin{subfigure}{0.49\textwidth}
        \centering
        \includegraphics[width=\linewidth]{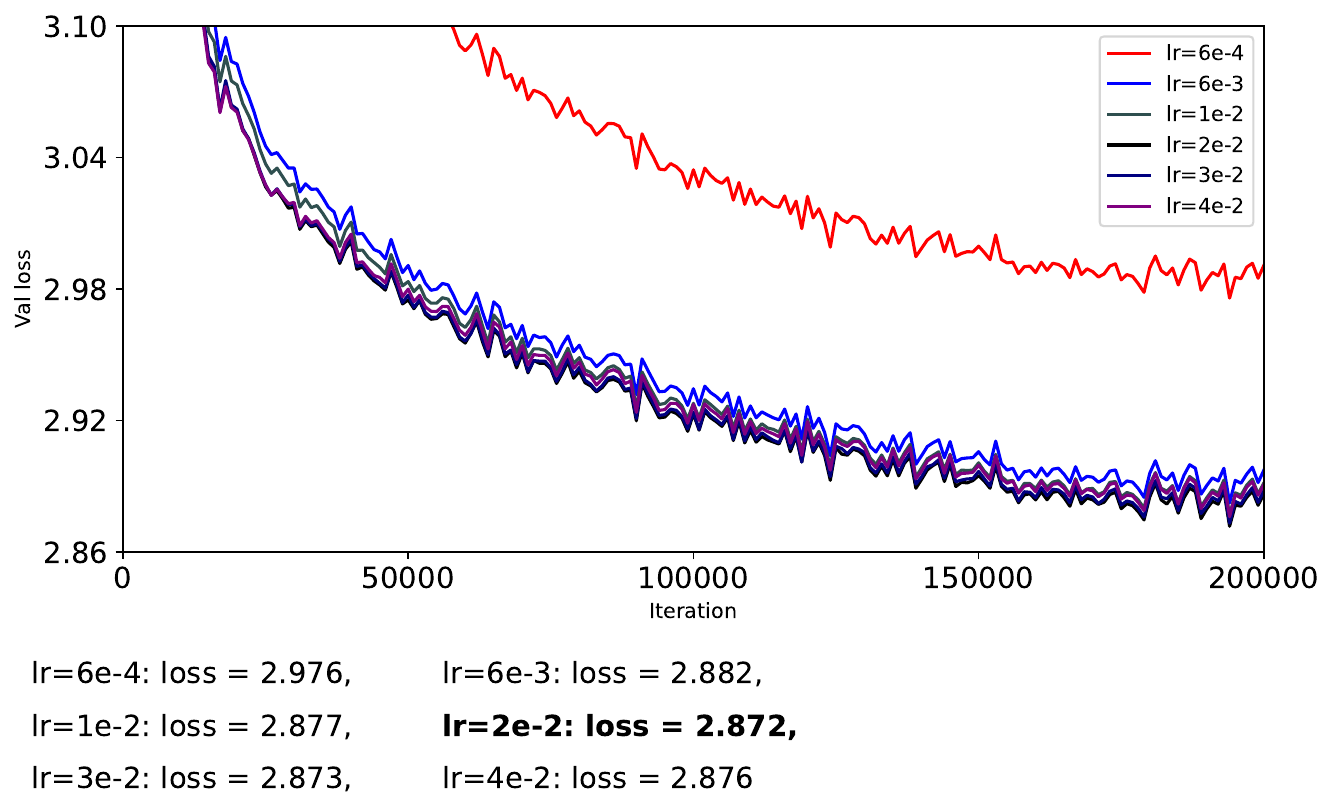}
        \caption{AdamQ ($\mU \mSigma^{\tfrac{1}{4}}\mV^{\top}$)}
        \label{fig:adam_p1_4}
    \end{subfigure}
    \begin{subfigure}{0.495\textwidth}
        \centering
        \includegraphics[width=\linewidth]{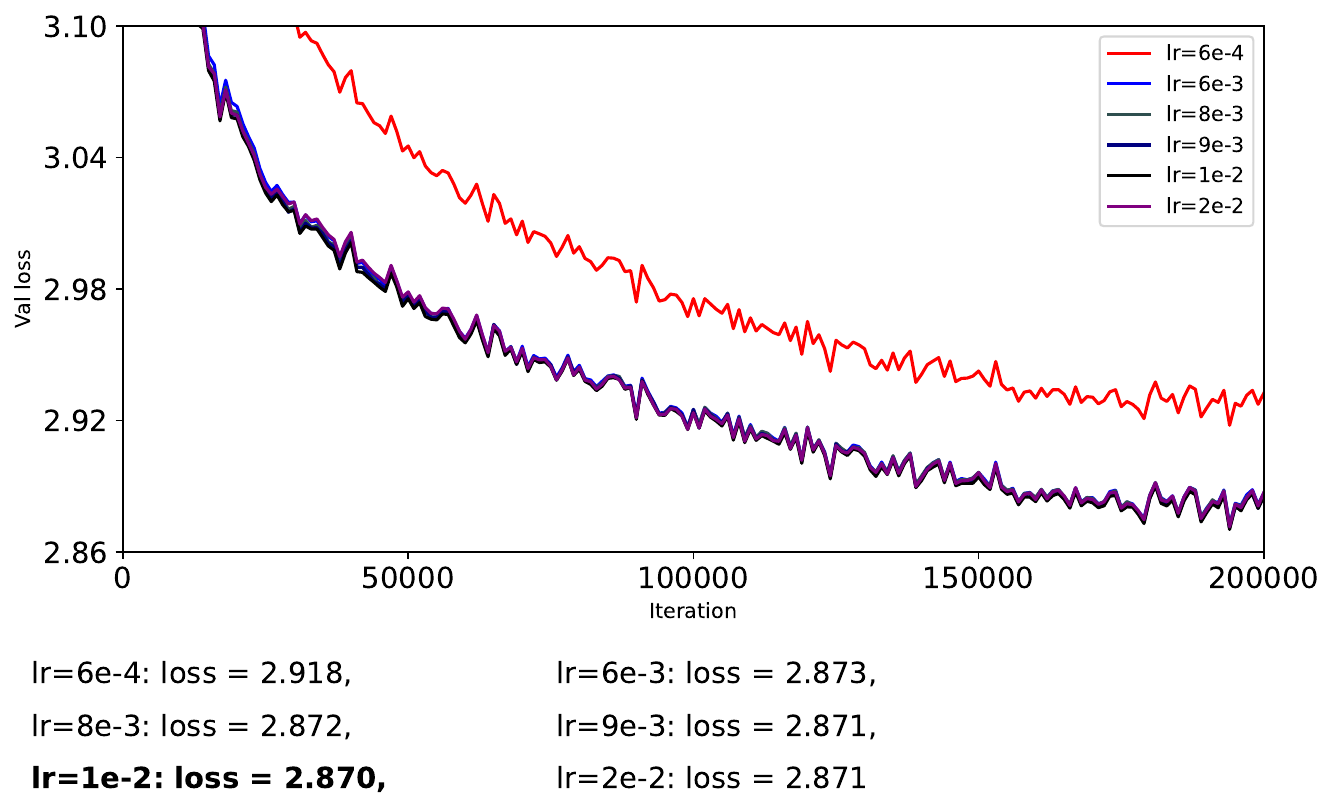}
        \caption{AdamS ($\mU \mSigma^{\tfrac{1}{2}}\mV^{\top}$)}
        \label{fig:adam_p1_2}
    \end{subfigure}
    \begin{subfigure}{0.495\textwidth}
        \centering
        \includegraphics[width=\linewidth]{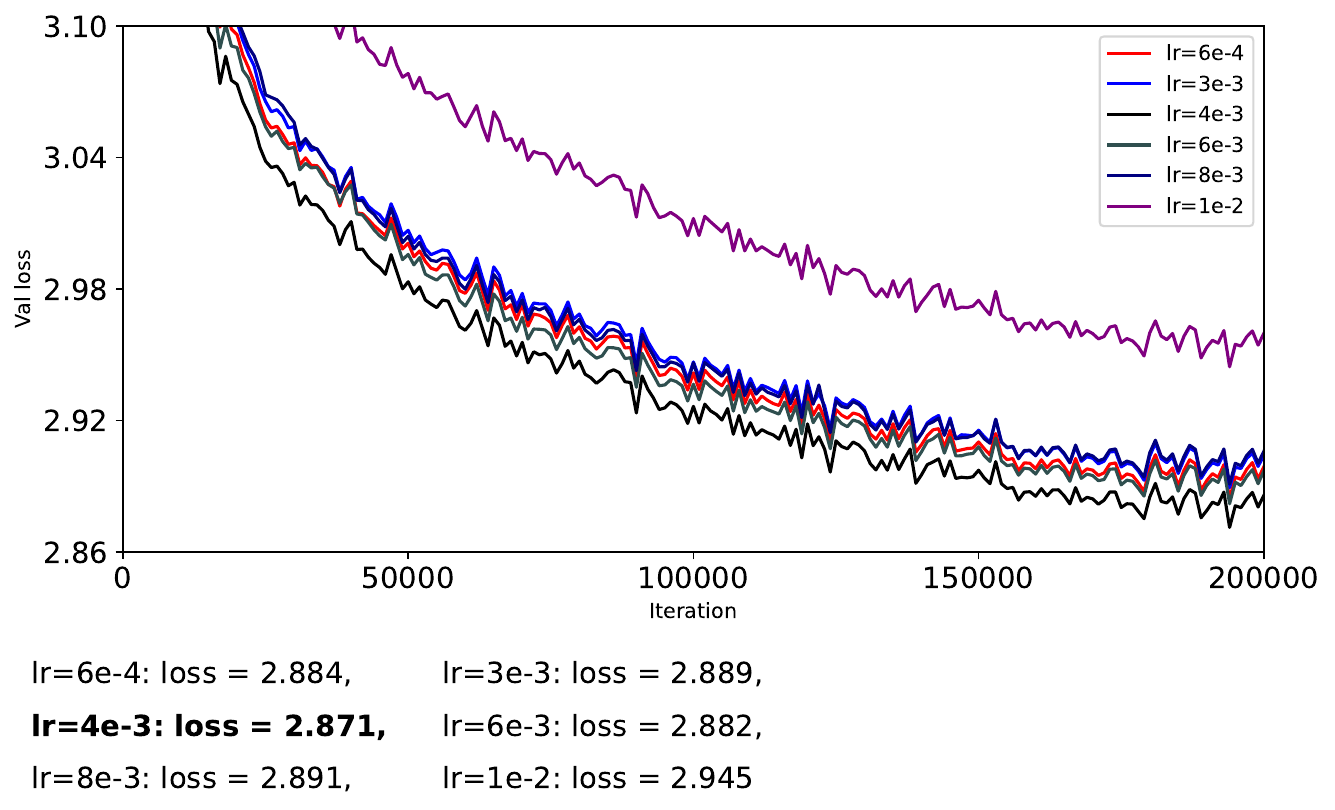}
        \caption{Adam ($\mU \mSigma^{1}\mV^{\top}$)}
        \label{fig:adam_p1}
    \end{subfigure}
    \caption{
        Overall comparison across four optimizers (AdamZ, AdamS, AdamQ and Adam) based on the \textbf{second-moment-normalized update $\mM_t \oslash \sqrt{\mV_t}$}.
        Each subfigure corresponds to a different optimizer.
    }
    \label{fig:adam_all}
\end{figure*}

\paragraph{Fine-grained search.}
After identifying a promising scale, we refine the search by evaluating a denser set of learning rates above and below the best coarse candidate. We stop refining once the selected learning rate performs better than its immediate neighbors in this local grid, indicating a stable local optimum within the explored range. For each optimizer, this refinement evaluates roughly ten candidates; for readability, we plot six representative runs in the figures.

\subsection{Effect of Spectral Exponent on Momentum Inputs}
\autoref{fig:msgd_all} reports representative tuning results when the input
$\mO_t^{\text{mom}}$ is used. We make three observations:
\begin{itemize}[leftmargin=*]
    \item \textit{Muon stabilizes momentum updates.}
    Muon (mSGDZ) is considerably more stable than mSGD across a wide learning-rate range.
    The best learning rate we find for Muon, $7\times 10^{-3}$, closely matches the value reported in prior benchmarking~\cite{wen_optimizer_wen2025fantastic}.
    \item \textit{Partial compression improves stability but not to Muon’s level.}
    The intermediate variants mSGDS and mSGDQ are more stable than mSGD, but remain less robust than Muon under aggressive learning rates.
    \item \textit{After stability is achieved, moderate compression can outperform flattening.}
    Among the four momentum-input variants, mSGDQ attains the strongest performance once it trains stably, exceeding Muon and 
    obtaining strong results at learning rates of \(4\times10^{-2}\) and \(5\times10^{-2}\). 
\end{itemize}

\subsection{Effect of Spectral Exponent on Normalized Inputs}
Similarly, \autoref{fig:adam_all} shows our selected results given rms-normalized momentum $\mO_t^{\text{rms}}$ as input. We observe that:
\begin{itemize}[leftmargin=*]
    \item \textit{Spectral transforms yield smaller gains when the input is already normalized.}
    Applying spectral compression to $\mO_t^{\text{rms}}$ produces limited improvements: AdamS can modestly broaden the stable learning-rate range, whereas stronger compression (AdamQ) offers little benefit and full flattening (AdamZ) degrades performance.
    \item \textit{All four variants behave similarly at their best settings.}
    The peak performance differences among Adam, AdamS, and AdamQ are modest, suggesting that elementwise normalization already controls much of the harmful anisotropy that spectral compression targets.
\end{itemize}

\subsection{Momentum vs. RMS-Normalized Inputs}
\begin{figure}[H]
    \centering
    \includegraphics[width=1.0\linewidth]{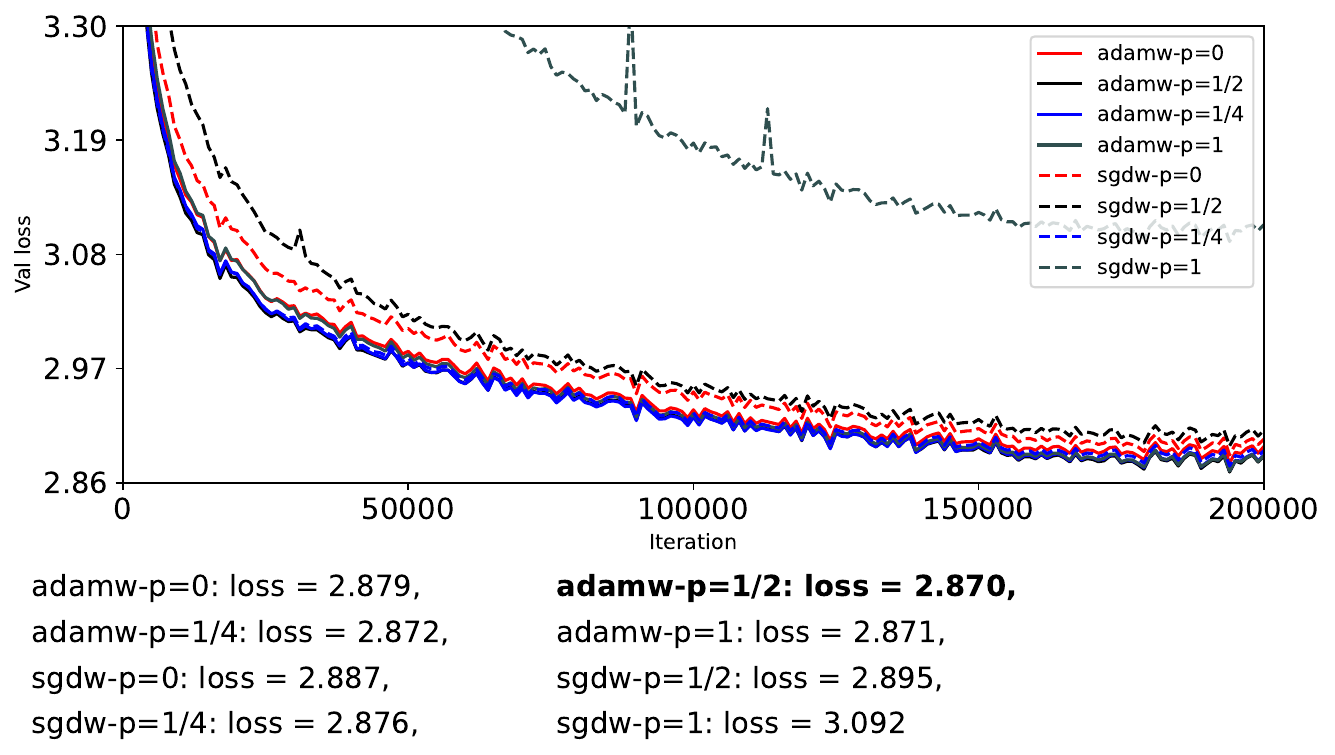}
    \caption{
    Comparison of four spectral exponents $p\in\{0,\tfrac14,\tfrac12,1\}$ applied to either the first-moment momentum $\mM_t$ or the RMS-normalized update $\mM_t \oslash \sqrt{\mV_t}$, each shown at its tuned learning rate.
    Dashed curves correspond to momentum-input variants, while solid curves correspond to RMS-normalized-input variants.
    }
    \label{fig:all_together}
\end{figure}

In~\autoref{fig:all_together}, we compare the best-tuned run from each of the eight configurations. Two trends are clear. First, switching the input from the momentum $\mO_t^{\text{mom}}$ to the RMS-normalized update $\mO_t^{\text{rms}}$ consistently improves both stability and final performance across spectral exponents. In particular, AdamZ (Muon-style orthogonalization applied to $\mO_t^{\text{rms}}$) is stronger than mSGDZ (the original Muon applied to $\mO_t^{\text{mom}}$). Second, within the RMS-normalized family, performance differences across $p$ are relatively small, with AdamS emerging as a strong and stable choice.

\section{Discussion: Our Understanding of Muon}
\label{sec:discussion}

The primary goal of this work is to evaluate when Muon-style orthogonalization (and, more broadly, spectral compression) is a reliable optimization strategy. Accordingly, we interpret Muon’s empirical behavior in \autoref{sec:experiments} through three honest conclusions.

\paragraph{\textbf{I. Muon exhibits no significant performance edge over AdamW.}}
In our controlled setting, Muon-style orthogonalization does not outperform Adam, whether applied to first-moment momentum or to RMS-normalized updates. This contrasts with prior reports of Muon converging faster than Adam~\cite{jordan_jordan2024muon}. We attribute the discrepancy primarily to the methodological differences noted in \autoref{sec:experiments:setup}. 
Broadly, we employ universal learning-rate decoupling, avoid auxiliary techniques like QK-Norm and QK-Clip, and disable weight decay.

\paragraph{\textbf{II. Muon is stable but aggressive.}}
On first-moment momentum inputs, Muon provides a clear stabilization effect where orthogonalization significantly improves robustness over mSGD across most learning rates (\autoref{fig:msgd_all}). This behavior aligns with Muon’s defining operation: given $\mG_t=\mU\mSigma\mV^\top$, Muon replaces the spectrum by $\mSigma^0$, producing the polar factor $\mU\mV^\top$. By discarding singular-value magnitudes, Muon prevents directions with extremely large singular values from dominating the update, mitigating ill-conditioning through \emph{direction-wise} spectral scale control.

However, this stabilization mechanism is also somewhat crude. Muon applies the same singular-value scaling to every direction, independent of curvature or signal strength. When some directions are legitimately larger than others, flattening them to unity can effectively reduce the step taken along informative directions. In this sense, Muon stabilizes by construction, but it does not selectively preserve meaningful anisotropy when such anisotropy is beneficial.

\paragraph{\textbf{III. Muon may magnify noisy gradient directions.}}
Muon’s biggest limitation follows from the same flattening: by mapping all nonzero singular values to $1$, it removes magnitude-based filtering. Directions corresponding to small singular values are no longer suppressed relative to dominant directions and can receive comparable update weight. This is especially problematic in the later stages of training where update matrices typically have low effective rank, and small-singular-value components may be primarily dominated by noise. 

This effect is consistent with our observations that aggressive Muon-like orthogonalization on RMS-normalized inputs does not outperform Adam and is actually often worse than partial compression. One plausible explanation is that RMS normalization already bounds and regularizes per-coordinate scales, so additional full-spectrum flattening primarily reweights the update toward directions that RMS normalization alone would keep small. 

\paragraph{Overall.} 
Together, our results suggest that spectral compression can provide genuine stabilization benefits, especially when the input update is \emph{not} already normalized. However, full orthogonalization ($p=0$) is not always desirable: it can under-step along genuinely informative dominant modes (crudeness) and over-step along low-signal modes (noise magnification). Hence, our experiments do not support orthogonalization as a universally superior replacement for modern RMS-normalized second-moment optimizers.

\section{Conclusion and Limitations}
\textbf{Conclusion.} This paper studies matrix-based optimizers from a spectral perspective. We interpret a range of optimization methods under a unified spectral framework.
Our results indicate that while spectral transformations like Muon have strong stabilizing properties, they cannot directly replace second-moment adaptive optimizers such as Adam.

\textbf{Limitations and future work.}
Our study omits weight decay to enable a controlled comparison and relies on a coupled Newton--Schulz iteration procedure that can incur nontrivial overhead. Investigating how to combine matrix-based updates with regularization techniques such as weight decay and improving the practical efficiency of the coupled Newton--Schulz algorithm are directions for future work.

\newpage

%\autoref{icml-historical}

\section*{Impact Statement}
``This paper presents work whose goal is to advance the field of Machine
Learning. There are many potential societal consequences of our work, none
which we feel must be specifically highlighted here.''

\bibliography{ref}

@article{sgd_robbins1951stochastic,
  title={A stochastic approximation method},
  author={Robbins, Herbert and Monro, Sutton},
  journal={The annals of mathematical statistics},
  pages={400--407},
  year={1951},
  publisher={JSTOR}
}

@article{lipsformer_qi2023lipsformer,
  title={LipsFormer: Introducing Lipschitz Continuity to Vision Transformers},
  author={Qi, Xianbiao and Wang, Jianan and Chen, Yihao and Shi, Yukai and Zhang, Lei},
  journal={arXiv preprint arXiv:2304.09856},
  year={2023}
}

@inproceedings{adafactor_shazeer2018adafactor,
  title={Adafactor: Adaptive learning rates with sublinear memory cost},
  author={Shazeer, Noam and Stern, Mitchell},
  booktitle={International Conference on Machine Learning},
  pages={4596--4604},
  year={2018},
  organization={PMLR}
}

@article{qi_qi2023understanding,
  title={Understanding optimization of deep learning via jacobian matrix and lipschitz constant},
  author={Qi, Xianbiao and Wang, Jianan and Zhang, Lei},
  journal={arXiv preprint arXiv:2306.09338},
  year={2023}
}

@article{layernorm_ba2016layer,
  title={Layer normalization},
  author={Ba, Jimmy Lei and Kiros, Jamie Ryan and Hinton, Geoffrey E},
  journal={arXiv preprint arXiv:1607.06450},
  year={2016}
}

@inproceedings{resnet_he2016deep,
  title={Deep residual learning for image recognition},
  author={He, Kaiming and Zhang, Xiangyu and Ren, Shaoqing and Sun, Jian},
  booktitle={Proceedings of the IEEE conference on computer vision and pattern recognition},
  pages={770--778},
  year={2016}
}

@inproceedings{msgd_nesterov1983method,
  title={A method for unconstrained convex minimization problem with the rate of convergence O (1/k\^{} 2)},
  author={Nesterov, Yurii},
  booktitle={Doklady an ussr},
  volume={269},
  pages={543--547},
  year={1983}
}

@article{svrg_johnson2013accelerating,
  title={Accelerating stochastic gradient descent using predictive variance reduction},
  author={Johnson, Rie and Zhang, Tong},
  journal={Advances in neural information processing systems},
  volume={26},
  year={2013}
}

@article{book_leon_bottou2018optimization,
  title={Optimization methods for large-scale machine learning},
  author={Bottou, L{\'e}on and Curtis, Frank E and Nocedal, Jorge},
  journal={SIAM review},
  volume={60},
  number={2},
  pages={223--311},
  year={2018},
  publisher={SIAM}
}

@inproceedings{signsgd_bernstein2018signsgd,
  title={signSGD: Compressed optimisation for non-convex problems},
  author={Bernstein, Jeremy and Wang, Yu-Xiang and Azizzadenesheli, Kamyar and Anandkumar, Animashree},
  booktitle={International Conference on Machine Learning},
  pages={560--569},
  year={2018},
  organization={PMLR}
}

@article{you_you2019large,
  title={Large batch optimization for deep learning: Training bert in 76 minutes},
  author={You, Yang and Li, Jing and Reddi, Sashank and Hseu, Jonathan and Kumar, Sanjiv and Bhojanapalli, Srinadh and Song, Xiaodan and Demmel, James and Keutzer, Kurt and Hsieh, Cho-Jui},
  journal={arXiv preprint arXiv:1904.00962},
  year={2019}
}

@article{wen_optimizer_wen2025fantastic,
  title={Fantastic pretraining optimizers and where to find them},
  author={Wen, Kaiyue and Hall, David and Ma, Tengyu and Liang, Percy},
  journal={arXiv preprint arXiv:2509.02046},
  year={2025}
}

@article{swishglu_shazeer2020glu,
  title={Glu variants improve transformer},
  author={Shazeer, Noam},
  journal={arXiv preprint arXiv:2002.05202},
  year={2020}
}

@article{adagrad_duchi2011adaptive,
  title={Adaptive subgradient methods for online learning and stochastic optimization.},
  author={Duchi, John and Hazan, Elad and Singer, Yoram},
  journal={Journal of machine learning research},
  volume={12},
  number={7},
  year={2011}
}

@article{lstm_hochreiter1997long,
  title={Long short-term memory},
  author={Hochreiter, Sepp and Schmidhuber, J{\"u}rgen},
  journal={Neural computation},
  volume={9},
  number={8},
  pages={1735--1780},
  year={1997},
  publisher={MIT press}
}

@article{cnn_lecun2002gradient,
  title={Gradient-based learning applied to document recognition},
  author={LeCun, Yann and Bottou, L{\'e}on and Bengio, Yoshua and Haffner, Patrick},
  journal={Proceedings of the IEEE},
  volume={86},
  number={11},
  pages={2278--2324},
  year={2002},
  publisher={Ieee}
}

@article{adam_adam2014method,
  title={A method for stochastic optimization},
  author={Adam, Kingma DP Ba J and others},
  journal={arXiv preprint arXiv:1412.6980},
  volume={1412},
  number={6},
  year={2014}
}

@article{rmsprop_hinton2012neural,
  title={Neural networks for machine learning lecture 6a overview of mini-batch gradient descent},
  author={Hinton, Geoffrey and Srivastava, Nitish and Swersky, Kevin},
  journal={Cited on},
  volume={14},
  number={8},
  pages={2},
  year={2012}
}

@article{suweijie_su2025isotropic,
  title={Isotropic Curvature Model for Understanding Deep Learning Optimization: Is Gradient Orthogonalization Optimal?},
  author={Su, Weijie},
  journal={arXiv preprint arXiv:2511.00674},
  year={2025}
}

@article{polargrad_lau2025polargrad,
  title={PolarGrad: A Class of Matrix-Gradient Optimizers from a Unifying Preconditioning Perspective},
  author={Lau, Tim Tsz-Kit and Long, Qi and Su, Weijie},
  journal={arXiv preprint arXiv:2505.21799},
  year={2025}
}

@article{what_frans2025really,
  title={What really matters in matrix-whitening optimizers?},
  author={Frans, Kevin and Abbeel, Pieter and Levine, Sergey},
  journal={arXiv preprint arXiv:2510.25000},
  year={2025}
}

@article{transformer_vaswani2017attention,
  title={Attention is all you need},
  author={Vaswani, Ashish and Shazeer, Noam and Parmar, Niki and Uszkoreit, Jakob and Jones, Llion and Gomez, Aidan N and Kaiser, {\L}ukasz and Polosukhin, Illia},
  journal={Advances in neural information processing systems},
  volume={30},
  year={2017}
}

@book{higham_higham2008functions,
  title={Functions of matrices: theory and computation},
  author={Higham, Nicholas J},
  year={2008},
  publisher={SIAM}
}

@misc{bernstein_bernstein2025deriving,
  author = {Jeremy Bernstein},
  title = {Deriving Muon},
  url = {https://jeremybernste.in/writing/deriving-muon},
  year = {2025}
}

@misc{newhouse_newhouse2025understanding,
  title = {Understanding Muon},
  author = {Laker Newhouse},
  url = {https://lakernewhouse.com/muon},
  year = {2025}
}

@article{bernstein_bernstein2025manifolds,
  author = {Jeremy Bernstein},
  title = {Modular Manifolds},
  journal = {Thinking Machines Lab: Connectionism},
  year = {2025},
  note = {https://thinkingmachines.ai/blog/modular-manifolds/},
  doi = {10.64434/tml.20250926}
}

@misc{nanogpt_Karpathy2022,
  author = {Andrej Karpathy},
  title = {\text{NanoGPT}},
  year = {2022},
  publisher = {GitHub},
  journal = {GitHub repository},
  howpublished = {\url{https://github.com/karpathy/nanoGPT}},
  commit = {325be85d9be8c81b436728a420e85796c57dba7e}
}

@misc{sam_buchanan2025mmuonadmm,
  author = {Buchanan, Sam},
  title = {A Faster Manifold Muon with {ADMM}},
  year = 2025,
  howpublished = {\url{https://sdbuchanan.com/blog/manifold-muon/}}
}

@misc{jordan_jordan2024muon,
  author       = {Keller Jordan and Yuchen Jin and Vlado Boza and Jiacheng You and
                  Franz Cesista and Laker Newhouse and Jeremy Bernstein},
  title        = {Muon: An optimizer for hidden layers in neural networks},
  year         = {2024},
  url          = {https://kellerjordan.github.io/posts/muon/}
}

@misc{adamuon_si2025adamuonadaptivemuonoptimizer,
      title={AdaMuon: Adaptive Muon Optimizer}, 
      author={Chongjie Si and Debing Zhang and Wei Shen},
      year={2025},
      eprint={2507.11005},
      archivePrefix={arXiv},
      primaryClass={cs.LG},
      url={https://arxiv.org/abs/2507.11005}, 
}

@misc{qwen_yang2025qwen3technicalreport,
      title={Qwen3 Technical Report}, 
      author={An Yang and Anfeng Li and Baosong Yang and Beichen Zhang and Binyuan Hui and Bo Zheng and Bowen Yu and Chang Gao and Chengen Huang and Chenxu Lv and Chujie Zheng and Dayiheng Liu and Fan Zhou and Fei Huang and Feng Hu and Hao Ge and Haoran Wei and Huan Lin and Jialong Tang and Jian Yang and Jianhong Tu and Jianwei Zhang and Jianxin Yang and Jiaxi Yang and Jing Zhou and Jingren Zhou and Junyang Lin and Kai Dang and Keqin Bao and Kexin Yang and Le Yu and Lianghao Deng and Mei Li and Mingfeng Xue and Mingze Li and Pei Zhang and Peng Wang and Qin Zhu and Rui Men and Ruize Gao and Shixuan Liu and Shuang Luo and Tianhao Li and Tianyi Tang and Wenbiao Yin and Xingzhang Ren and Xinyu Wang and Xinyu Zhang and Xuancheng Ren and Yang Fan and Yang Su and Yichang Zhang and Yinger Zhang and Yu Wan and Yuqiong Liu and Zekun Wang and Zeyu Cui and Zhenru Zhang and Zhipeng Zhou and Zihan Qiu},
      year={2025},
      eprint={2505.09388},
      archivePrefix={arXiv},
      primaryClass={cs.CL},
      url={https://arxiv.org/abs/2505.09388}, 
}

@article{newton_schulz_schulz1933iterative,
  title={Iterative berechung der reziproken matrix},
  author={Schulz, G{\"u}nther},
  journal={ZAMM-Journal of Applied Mathematics and Mechanics/Zeitschrift f{\"u}r Angewandte Mathematik und Mechanik},
  volume={13},
  number={1},
  pages={57--59},
  year={1933},
  publisher={Wiley Online Library}
}

@article{kimi_kimiteam2025kimik2openagentic,
  title={Kimi k2: Open agentic intelligence},
  author={Team, Kimi and Bai, Yifan and Bao, Yiping and Chen, Guanduo and Chen, Jiahao and Chen, Ningxin and Chen, Ruijue and Chen, Yanru and Chen, Yuankun and Chen, Yutian and others},
  journal={arXiv preprint arXiv:2507.20534},
  year={2025}
}

@article{deepseek_deepseekai2025deepseekr1incentivizingreasoningcapability,
  title={Deepseek-r1: Incentivizing reasoning capability in llms via reinforcement learning},
  author={Guo, Daya and Yang, Dejian and Zhang, Haowei and Song, Junxiao and Zhang, Ruoyu and Xu, Runxin and Zhu, Qihao and Ma, Shirong and Wang, Peiyi and Bi, Xiao and others},
  journal={arXiv preprint arXiv:2501.12948},
  year={2025}
}

@misc{adamw_loshchilov2019decoupled,
      title={Decoupled Weight Decay Regularization}, 
      author={Ilya Loshchilov and Frank Hutter},
      year={2019},
      eprint={1711.05101},
      archivePrefix={arXiv},
      primaryClass={cs.LG},
      url={https://arxiv.org/abs/1711.05101}, 
}

@misc{lion_chen2023symbolic,
      title={Symbolic Discovery of Optimization Algorithms}, 
      author={Xiangning Chen and Chen Liang and Da Huang and Esteban Real and Kaiyuan Wang and Yao Liu and Hieu Pham and Xuanyi Dong and Thang Luong and Cho-Jui Hsieh and Yifeng Lu and Quoc V. Le},
      year={2023},
      eprint={2302.06675},
      archivePrefix={arXiv},
      primaryClass={cs.LG},
      url={https://arxiv.org/abs/2302.06675}, 
}

@misc{mars_yuan2025mars,
      title={MARS: Unleashing the Power of Variance Reduction for Training Large Models}, 
      author={Huizhuo Yuan and Yifeng Liu and Shuang Wu and Xun Zhou and Quanquan Gu},
      year={2025},
      eprint={2411.10438},
      archivePrefix={arXiv},
      primaryClass={cs.LG},
      url={https://arxiv.org/abs/2411.10438}, 
}

@misc{adam_mini_zhang2025adamminiusefewerlearning,
      title={Adam-mini: Use Fewer Learning Rates To Gain More}, 
      author={Yushun Zhang and Congliang Chen and Ziniu Li and Tian Ding and Chenwei Wu and Diederik P. Kingma and Yinyu Ye and Zhi-Quan Luo and Ruoyu Sun},
      year={2025},
      eprint={2406.16793},
      archivePrefix={arXiv},
      primaryClass={cs.LG},
      url={https://arxiv.org/abs/2406.16793}, 
}

@article{llama3_grattafiori2024llama3herdmodels,
  title={The llama 3 herd of models},
  author={Dubey, Abhimanyu and Jauhri, Abhinav and Pandey, Abhinav and Kadian, Abhishek and Al-Dahle, Ahmad and Letman, Aiesha and Mathur, Akhil and Schelten, Alan and Yang, Amy and Fan, Angela and others},
  journal={arXiv e-prints},
  pages={arXiv--2407},
  year={2024}
}

@article{llama2_touvron2023llama2openfoundation,
  title={Llama 2: Open foundation and fine-tuned chat models},
  author={Touvron, Hugo and Martin, Louis and Stone, Kevin and Albert, Peter and Almahairi, Amjad and Babaei, Yasmine and Bashlykov, Nikolay and Batra, Soumya and Bhargava, Prajjwal and Bhosale, Shruti and others},
  journal={arXiv preprint arXiv:2307.09288},
  year={2023}
}

@article{llama_touvron2023llamaopenefficientfoundation,
	title={Llama: Open and efficient foundation language models},
	author={Touvron, Hugo and Lavril, Thibaut and Izacard, Gautier and Martinet, Xavier and Lachaux, Marie-Anne and Lacroix, Timoth{\'e}e and Rozi{\`e}re, Baptiste and Goyal, Naman and Hambro, Eric and Azhar, Faisal and others},
	journal={arXiv preprint arXiv:2302.13971},
	year={2023}
}

@misc{sophia_liu2024sophiascalablestochasticsecondorder,
      title={Sophia: A Scalable Stochastic Second-order Optimizer for Language Model Pre-training}, 
      author={Hong Liu and Zhiyuan Li and David Hall and Percy Liang and Tengyu Ma},
      year={2024},
      eprint={2305.14342},
      archivePrefix={arXiv},
      primaryClass={cs.LG},
      url={https://arxiv.org/abs/2305.14342}, 
}

@misc{cautious_liang2025cautious,
      title={Cautious Optimizers: Improving Training with One Line of Code}, 
      author={Kaizhao Liang and Lizhang Chen and Bo Liu and Qiang Liu},
      year={2025},
      eprint={2411.16085},
      archivePrefix={arXiv},
      primaryClass={cs.LG},
      url={https://arxiv.org/abs/2411.16085}, 
}

@misc{adan_xie2024adanadaptivenesterovmomentum,
      title={Adan: Adaptive Nesterov Momentum Algorithm for Faster Optimizing Deep Models}, 
      author={Xingyu Xie and Pan Zhou and Huan Li and Zhouchen Lin and Shuicheng Yan},
      year={2024},
      eprint={2208.06677},
      archivePrefix={arXiv},
      primaryClass={cs.LG},
      url={https://arxiv.org/abs/2208.06677}, 
}

@inproceedings{kfac_martens2015optimizing,
  title={Optimizing neural networks with kronecker-factored approximate curvature},
  author={Martens, James and Grosse, Roger},
  booktitle={International conference on machine learning},
  pages={2408--2417},
  year={2015},
  organization={PMLR}
}

@inproceedings{shampoo_gupta2018shampoo,
  title={Shampoo: Preconditioned stochastic tensor optimization},
  author={Gupta, Vineet and Koren, Tomer and Singer, Yoram},
  booktitle={International Conference on Machine Learning},
  pages={1842--1850},
  year={2018},
  organization={PMLR}
}

@article{higham_higham1986computing_polar_decomp,
  title={Computing the polar decomposition—with applications},
  author={Higham, Nicholas J},
  journal={SIAM Journal on Scientific and Statistical Computing},
  volume={7},
  number={4},
  pages={1160--1174},
  year={1986},
  publisher={SIAM}
}

@article{higham_higham1987computing_square_root,
  title={Computing real square roots of a real matrix},
  author={Higham, Nicholas J},
  journal={Linear Algebra and its applications},
  volume={88},
  pages={405--430},
  year={1987},
  publisher={Elsevier}
}

@techreport{jan_zur_matrix_functions,
  author       = {Zur, Jan},
  title        = {Matrix Functions},
  institution  = {Technische Universität Berlin},
  year         = {2025},
  url          = {https://www.tu.berlin/fileadmin/www/40000110/Dauer-Wimis/Jan_Zur/matrix_functions_zur.pdf},
  note         = {Lecture note}
}

@book{matrix_analysis_horn2012matrix,
  title={Matrix analysis},
  author={Horn, Roger A and Johnson, Charles R},
  year={2012},
  publisher={Cambridge university press}
}
\bibliographystyle{icml2026}

%%%%%%%%%%%%%%%%%%%%%%%%%   %%%%%%%%%%%%%%%%%%%%%%%%%%%%%%%%%%%%%%%%%%%%%%%%%%%%%%
%%%%%%%%%%%%%%%%%%%%%%%%%%%%%%%%%%%%%%%%%%%%%%%%%%%%%%%%%%%%%%%%%%%%%%%%%%%%%%%

% APPENDIX
%%%%%%%%%%%%%%%%%%%%%%%%%%%%%%%%%%%%%%%%%%%%%%%%%%%%%%%%%%%%%%%%%%%%%%%%%%%%%%%
%%%%%%%%%%%%%%%%%%%%%%%%%%%%%%%%%%%%%%%%%%%%%%%%%%%%%%%%%%%%%%%%%%%%%%%%%%%%%%%
\newpage
\appendix
\onecolumn

Almost all derivations are about functions of matrices, all these derivatinos can be found in the following materials~\cite{higham_higham1986computing_polar_decomp, higham_higham1987computing_square_root, higham_higham2008functions, jan_zur_matrix_functions, matrix_analysis_horn2012matrix}.

\section{Derivation of $\mU\mV^{\top}$ based on Newton-Schulz method}\label{sec:appendix:muon}

Let the singular value decomposition of a matrix $\mM\in\mathbb{R}^{m\times n}$ be
\[
\mM = \mU \mS \mV^\top,
\]
where
$\mU\in\mathbb{R}^{m\times r}, 
\mV\in\mathbb{R}^{n\times r}, 
\mS=\operatorname{Diag}(\sigma_1,\dots,\sigma_r), \sigma_i>0,$
and $\mU,\mV$ have orthonormal columns. Let us assume $m\leq n$

Our goal is to compute the orthogonal (polar) factor $\mU\mV^\top$
\emph{without explicitly performing an SVD}, using the Newton--Schulz iteration.

% ============================================================
\paragraph{Step 1: Express $\mU\mV^\top$ as a matrix function of $\mM$.}

We start from quantities that can be formed directly from $\mM$.
Consider
$\mM^\top \mM \in \mathbb{R}^{n\times n}.$
Substituting the SVD of $\mM$ gives
\[
\mM^\top \mM = (\mU\mS\mV^\top)^\top (\mU\mS\mV^\top) = \mV \mS \mU^\top \mU \mS \mV^\top = \mV \mS^2 \mV^\top,
\]
where we used $\mU^\top \mU = \mI_r$.

Hence, the inverse square root of $\mM^\top \mM$ (on its rank-$r$ subspace) is
\[
(\mM^\top \mM)^{-1/2} = \mV \mS^{-1} \mV^\top.
\]
Left-multiplying by $\mM$ yields
\[
\mM(\mM^\top \mM)^{-1/2} = (\mU\mS\mV^\top)(\mV\mS^{-1}\mV^\top) = \mU(\mS\mS^{-1})\mV^\top = \mU\mV^\top.
\]

This identity shows that $\mU\mV^\top$ can be obtained by removing the singular-value scaling of $\mM$ via $(\mM^\top \mM)^{-1/2}$.

% ============================================================
\paragraph{Step 2: Reduction to an inverse square root problem.}

Define
\[
\mA := \mM^\top \mM \succeq 0.
\]
The problem is now reduced to computing
$\mA^{-1/2}$
using only matrix multiplication, without SVD.

% ============================================================
\paragraph{Step 3: Newton--Schulz iteration for the inverse square root.}
Let us consider a scalar $a>0$, consider the problem of computing
$z = a^{-1/2}.$
This is equivalent to solving
$
f(z) = \frac{1}{z^2} - a = 0.
$
Applying Newton's method gives
\begin{equation*}
    z^{+}
= z - \frac{f(z)}{f'(z)}
= z - \frac{z^{-2}-a}{-2z^{-3}} = \frac12\, z(3 - a z^2).
\end{equation*}
This is the scalar Newton--Schulz update.

Replacing the scalar $a$ by a matrix $\mA\succeq 0$, the scalar $z$ by a matrix $\mZ$, and scalar multiplication by matrix multiplication yields the matrix Newton--Schulz iteration
\[
\mZ_{k+1}
=
\frac12\, \mZ_k\bigl(3\mI - \mA \mZ_k^2\bigr).
\]
This iteration is equal to \autoref{eq:ns_muon} in the main body. It
involves only matrix multiplication, addition, and scalar scaling.
After sufficient iterations,
\[
\mZ_k \approx \mA^{-1/2}.
\]

% ============================================================
\paragraph{Step 4: Recovering $\mU\mV^\top$.}
Using the identity
$\mU\mV^\top = \mM \mA^{-1/2}, \mA=\mM^\top\mM$,
and the approximation $\mZ_k\approx \mA^{-1/2}$, we obtain
\[
\mU\mV^\top \approx \mM \mZ_k.
\]

\ 

\ 

\

\section{Newton-Schulz method used by~\citet{jordan_jordan2024muon}}
\label{appendix:NS_by_jordan}
\begin{algorithm}[ht]
\caption{Newton-Schulz to solve $\mU\mV^{\top}$ \text{where} $\mM = \mU\mSigma \mV^{\top}$}
\label{alg:ns5_muon}
\begin{algorithmic}[1]
\REQUIRE  $\mM$, $K$
\ENSURE $\mU\mV^{\top}$

\STATE \textbf{Coefficients:} $(a,b,c) \leftarrow (3.4445,\,-4.7750,\,2.0315)$

\STATE $\mZ_0 \leftarrow \mM$
\STATE $\alpha \leftarrow \|\mZ_0\|_F$
\STATE $\mZ_0 \leftarrow \mZ_0 / \alpha,\ k \leftarrow  0$
\WHILE{$k < K$}
    \STATE $\mZ_{k+1} \leftarrow a\,\mZ_k + b\, \mZ_k \mZ_k^{\top} \mZ_k + c\,(\mZ_k \mZ_k^{\top})^2 \mZ_k$
    \STATE $k \leftarrow k + 1$
\ENDWHILE

\STATE \textbf{return} $\mZ_K$
\end{algorithmic}
\end{algorithm}

\ 

\ 

\

\section{Derive $\mX^{\frac{1}{2}}$ based on Coupled Newton-Schulz method}

We aim to compute the matrix square root $\mX^{1/2}$.
Let the unknown matrix $\mY$ satisfy
\[
f(\mY) := \mY^2 - \mX = 0 .
\]
This defines a matrix equation
$f : \mathbb{C}^{n\times n} \longrightarrow \mathbb{C}^{n\times n}.$

\paragraph{Fr\'echet Derivative.}
For an arbitrary perturbation $\mH$, we have
\[
f(\mY + \mH)
= (\mY + \mH)^2 - \mX
= \mY^2 - \mX + (\mY\mH + \mH\mY) + \mH^2 .
\]
Therefore, the Fr\'echet derivative of $f$ at $\mY$ is
\[
\mD f(\mY)[\mH] = \mY\mH + \mH\mY .
\]

\paragraph{Definition of the Newton Step.}
The Newton increment $\Delta \mY$ at $\mY$ is defined as the solution of
\[
\mD f(\mY)[\Delta \mY] = -f(\mY),
\]
that is, the Sylvester equation
\begin{equation}
\mY \Delta \mY + \Delta \mY\, \mY = -(\mY^2 - \mX).
\label{eq:newton-sylvester}
\end{equation}
The update is then given by
\[
\mY^+ = \mY + \Delta \mY .
\]

Up to this point, the derivation corresponds to the \emph{exact matrix Newton method},
without any approximation.

% ============================================================

\paragraph{A Coupled Newton System.}
The main difficulty of \eqref{eq:newton-sylvester} is that it requires solving a
Sylvester equation.
A classical technique is to couple the unknown square root with its inverse,
thereby converting the Sylvester solve into matrix multiplications.

Let $\mZ$ approximate $\mY^{-1}$ (and ultimately $\mX^{-1/2}$).
Consider the coupled system
\begin{equation}
\begin{cases}
F_1(\mY,\mZ) := \mY^2 - \mX = 0, \\
F_2(\mY,\mZ) := \mZ \mY - \mI = 0 .
\end{cases}
\label{eq:coupled-system}
\end{equation}
The exact solution is
\[
(\mY_\star, \mZ_\star) = (\mX^{1/2}, \mX^{-1/2}).
\]

\paragraph{Fr\'echet Derivative of the Coupled System}

For perturbations $(\Delta \mY, \Delta \mZ)$:
\begin{itemize}
\item For $F_1$,
\[
\mD F_1(\mY,\mZ)[\Delta \mY, \Delta \mZ]
= \mY \Delta \mY + \Delta \mY\, \mY .
\]

\item For $F_2$,
\[
(\mZ + \Delta \mZ)(\mY + \Delta \mY) - \mI
= (\mZ\mY - \mI)
+ (\mZ \Delta \mY + \Delta \mZ\, \mY)
+ (\Delta \mZ)(\Delta \mY),
\]
hence
\[
\mD F_2(\mY,\mZ)[\Delta \mY, \Delta \mZ]
= \mZ \Delta \mY + \Delta \mZ\, \mY .
\]
\end{itemize}

\paragraph{Newton Linear System (Exact)}

The Newton step $(\Delta \mY, \Delta \mZ)$ is defined by
\[
\mD \mF(\mY,\mZ)[\Delta \mY, \Delta \mZ]
= -\mF(\mY,\mZ),
\]
that is,
\begin{equation}
\begin{aligned}
\mY \Delta \mY + \Delta \mY\, \mY &= -(\mY^2 - \mX),\\
\mZ \Delta \mY + \Delta \mZ\, \mY &= -(\mZ \mY - \mI).
\end{aligned}
\label{eq:coupled-newton-linear}
\end{equation}
This system is still an \emph{exact Newton system}.

% ============================================================

\paragraph{A Closed-Form Newton Step via a Symmetric Multiplicative Ansatz}

We adopt the symmetric multiplicative ansatz
\begin{equation}
\mY^+ = \mY \mQ,
\qquad
\mZ^+ = \mQ \mZ ,
\label{eq:symmetric-update}
\end{equation}
where $\mQ = \mQ(\mY,\mZ)$ is to be determined.

Define
\begin{equation*}
\mR := \mZ \mY .
\label{eq:W-def}
\end{equation*}
Then
\begin{equation*}
\mR^+ := \mZ^+ \mY^+
= (\mQ \mZ)(\mY \mQ)
= \mQ (\mZ \mY) \mQ
= \mQ \mR \mQ .
\label{eq:W-update}
\end{equation*}

Hence, choosing $\mQ$ reduces to the problem:
given $\mR$, construct $\mQ$ such that $\mQ \mR \mQ \approx \mI$.

% ============================================================

\paragraph{Deriving $\mQ = \tfrac12(3\mI - \mZ\mY)$ from Newton's Principle.} To approximate $\mR^{-1/2}$, ideally,
\[
\mQ = \mR^{-1/2}
\quad \Rightarrow \quad
\mQ \mR \mQ = \mI .
\]

Consider the matrix function
\[
\phi(\mR) := \mR^{-1/2}.
\]
When $\mR$ is close to the identity, write
\[
\mR = \mI + \mE,
\qquad \|\mE\| \ll 1 .
\]
Accoroding to the Taylor expansion, we have
\[
(\mI + \mE)^{-1/2}
= \mI - \tfrac12 \mE + O(\|\mE\|^2).
\]
Substituting $\mE = \mR - \mI$, we obtain
\begin{equation*}
\mR^{-1/2}
\approx \mI - \tfrac12(\mR - \mI)
= \tfrac12(3\mI - \mR).
\label{eq:first-order-inv-sqrt}
\end{equation*}

Thus define
\begin{equation}
\mQ := \tfrac12(3\mI - \mR)
= \tfrac12(3\mI - \mZ \mY).
\label{eq:M-def}
\end{equation}

\paragraph{Deriving $\mQ = \tfrac12(3\mI - \mZ\mY)$ from Newton's Principle}{Coupled Newton--Schulz Iteration.}

Substituting \eqref{eq:M-def} into \eqref{eq:symmetric-update}, we obtain
\begin{equation*}
\begin{aligned}
\mY^+ &= \tfrac12\, \mY (3\mI - \mZ \mY),\\
\mZ^+ &= \tfrac12\, (3\mI - \mZ \mY)\, \mZ .
\end{aligned}
\label{eq:coupled-NS-update}
\end{equation*}

Write it as an iteration equation, we have
\begin{equation}
\begin{aligned}
\mY_{k+1} &= \tfrac12\, \mY_k (3\mI - \mZ_k \mY_k),
\qquad \mY_0 = \mX,\\
\mZ_{k+1} &= \tfrac12\, (3\mI - \mZ_k \mY_k)\, \mZ_k,
\qquad \mZ_0 = \mI .
\end{aligned}
\label{eq:coupled-NS-iteration}
\end{equation}

This is the \emph{coupled Newton--Schulz iteration}, which simultaneously drives
\[
\mY_k \to \mX^{1/2},
\qquad
\mZ_k \to \mX^{-1/2}.
\]

\ 

\ 

\

\section{Derive $\mX^{\frac{1}{4}}$ based on Coupled Newton-Schulz method}

\begin{algorithm}
\caption{Two-times Coupled Newton-Schulz to solve $\mX^{\frac{1}{4}}$ and $\mX^{\frac{-1}{4}}$}
\label{alg:couple_ns_1_4}
\begin{algorithmic}[1]
\REQUIRE $\mX, K$
\ENSURE $\mX^{\frac{1}{4}}, \mX^{\frac{-1}{4}}$
\STATE $\mY_0 \leftarrow \mX$, $\mZ_0 \leftarrow \mI$
\STATE $\alpha \leftarrow \|\mX\|_F$
\STATE  $\mY_0 \leftarrow \frac{\mY_0}{\alpha}$, $k \leftarrow  0$
\WHILE{$k < K$}
    \STATE $\mT_k \leftarrow 3\mI - \mZ_k \mY_k$
    \STATE $\mY_{k+1} \leftarrow \tfrac{1}{2}\, \mY_k \mT_k$
    \STATE $\mZ_{k+1} \leftarrow \tfrac{1}{2}\, \mT_k \mZ_k$
    \STATE $k \leftarrow k + 1$
\ENDWHILE

\STATE $\mY_0 \leftarrow \sqrt{\alpha} \mY_K$, $\mZ_0 \leftarrow \mI$
\STATE $\beta \leftarrow \|\mY_0\|_F$
\STATE  $\mY_0 \leftarrow \frac{\mY_0}{\beta}$, $k \leftarrow  0$
\WHILE{$k < K$}
    \STATE $\mT_k \leftarrow 3\mI - \mZ_k \mY_k$
    \STATE $\mY_{k+1} \leftarrow \tfrac{1}{2}\, \mY_k \mT_k$
    \STATE $\mZ_{k+1} \leftarrow \tfrac{1}{2}\, \mT_k \mZ_k$
    \STATE $k \leftarrow k + 1$
\ENDWHILE
\STATE \textbf{return} $\sqrt{\beta}\,\mY_K,\; \frac{1}{\sqrt{\beta}}\,\mZ_K$
%\STATE \textbf{return} $\mB_K$, $\mC_K$
%$\frac{\partial l}{\partial \mA} \leftarrow \frac{1}{2} \mC_k$
\end{algorithmic}
\end{algorithm}

\newpage
\section{Detailed results for different settings}

\begin{table}[h!]
\centering
\caption{\textbf{Hyperparameter sweeps for Muon.}}
\label{tab:msgd_sweeps}
\setlength{\tabcolsep}{10pt} % column padding
\renewcommand{\arraystretch}{1.15} % row height

\begin{tabular}{llccc}
\toprule
\textbf{Origin} &
\textbf{$\Psi_p(\mO^{\text{mom}})$} &
\textbf{Learning Rate} &
\textbf{Weight Decay} & 
\textbf{Val Loss} \\
\midrule
First order  & $\mU \mSigma^{1}\mV^{\top}$   & 1e-1 & 0 & 3.649 \\
First order  & $\mU \mSigma^{1}\mV^{\top}$   & 5e-1 & 0 & 3.226 \\
First order  & $\mU \mSigma^{1}\mV^{\top}$   & 1.0 & 0 & 3.147 \\
First order  & $\mU \mSigma^{1}\mV^{\top}$   & 2.0 & 0 & \textbf{3.092} \\
First order  & $\mU \mSigma^{1}\mV^{\top}$   & 5.0 & 0 & 3.092 \\
First order  & $\mU \mSigma^{1}\mV^{\top}$   & 10.0 & 0 & 3.164 \\
\hline 
First order  & $\mU \mSigma^{\tfrac{1}{2}}\mV^{\top}$   & 1e-2 & 0 & 3.054 \\
First order  & $\mU \mSigma^{\tfrac{1}{2}}\mV^{\top}$   & 3e-2 & 0 & 2.955 \\
First order  & $\mU \mSigma^{\tfrac{1}{2}}\mV^{\top}$   & 1e-1 & 0 & 2.908 \\
First order  & $\mU \mSigma^{\tfrac{1}{2}}\mV^{\top}$   & 2e-1 & 0 & \textbf{2.895} \\
First order  & $\mU \mSigma^{\tfrac{1}{2}}\mV^{\top}$   & 1.25e-1 & 0 & 2.904 \\
First order  & $\mU \mSigma^{\tfrac{1}{2}}\mV^{\top}$   & 3e-1 & 0 & inf \\
\hline 

First order  & $\mU \mSigma^{\tfrac{1}{4}}\mV^{\top}$   & 3e-3 & 0 & 3.030 \\
First order  & $\mU \mSigma^{\tfrac{1}{4}}\mV^{\top}$   & 6e-3 & 0 & 2.969 \\
First order  & $\mU \mSigma^{\tfrac{1}{4}}\mV^{\top}$   & 1e-2 & 0 & 2.938 \\
First order  & $\mU \mSigma^{\tfrac{1}{4}}\mV^{\top}$   & 3e-2 & 0 & 2.904 \\
First order  & $\mU \mSigma^{\tfrac{1}{4}}\mV^{\top}$   & 4e-2 & 0 & \textbf{2.876} \\
First order  & $\mU \mSigma^{\tfrac{1}{4}}\mV^{\top}$   & 5e-2 & 0 & 2.879 \\
\hline 

First order  & $\mU \mSigma^{0}\mV^{\top}$   & 3e-3 & 0 & 2.891 \\
First order  & $\mU \mSigma^{0}\mV^{\top}$   & 6e-3 & 0 & 2.888 \\
First order  & $\mU \mSigma^{0}\mV^{\top}$   & 7e-3 & 0 & \textbf{2.887} \\
First order  & $\mU \mSigma^{0}\mV^{\top}$   & 8e-3 & 0 & 2.889 \\
First order  & $\mU \mSigma^{0}\mV^{\top}$   & 9e-3 & 0 & 2.891 \\
First order  & $\mU \mSigma^{0}\mV^{\top}$   & 1e-2 & 0 & 2.891\\
\hline 
\bottomrule
\end{tabular}
\end{table}

\begin{table}[h!]
\centering
\caption{\textbf{Hyperparameter sweeps for Muon.}}
\label{tab:adam_sweeps}
\setlength{\tabcolsep}{10pt} % column padding
\renewcommand{\arraystretch}{1.15} % row height

\begin{tabular}{llccc}
\toprule
\textbf{Origin} &
\textbf{$\Psi_p(\mO^{\text{rms}})$} &
\textbf{Learning Rate} &
\textbf{Weight Decay} & 
\textbf{Val Loss} \\
\midrule
Second order  & $\mU \mSigma^{1}\mV^{\top}$   & 6e-4 & 0 & 2.884 \\
Second order  & $\mU \mSigma^{1}\mV^{\top}$   & 3e-3 & 0 & 2.889 \\
Second order  & $\mU \mSigma^{1}\mV^{\top}$   & 4e-3 & 0 & \textbf{2.871} \\
Second order  & $\mU \mSigma^{1}\mV^{\top}$   & 6e-3 & 0 & 2.882 \\
Second order  & $\mU \mSigma^{1}\mV^{\top}$   & 8e-3 & 0 & 2.891 \\
Second order  & $\mU \mSigma^{1}\mV^{\top}$   & 1e-2 & 0 & 2.945 \\
\hline 
Second order  & $\mU \mSigma^{\tfrac{1}{2}}\mV^{\top}$   & 6e-4 & 0 & 2.918 \\
Second order  & $\mU \mSigma^{\tfrac{1}{2}}\mV^{\top}$   & 6e-3 & 0 & 2.873 \\
Second order  & $\mU \mSigma^{\tfrac{1}{2}}\mV^{\top}$   & 8e-3 & 0 & 2.872 \\
Second order  & $\mU \mSigma^{\tfrac{1}{2}}\mV^{\top}$   & 9e-3 & 0 & 2.871 \\
Second order  & $\mU \mSigma^{\tfrac{1}{2}}\mV^{\top}$   & 1e-2 & 0 & \textbf{2.870} \\
Second order  & $\mU \mSigma^{\tfrac{1}{2}}\mV^{\top}$   & 2e-2 & 0 & 2.871 \\
\hline 

Second order  & $\mU \mSigma^{\tfrac{1}{4}}\mV^{\top}$   & 6e-4 & 0 & 2.976 \\
Second order  & $\mU \mSigma^{\tfrac{1}{4}}\mV^{\top}$   & 6e-3 & 0 & 2.882 \\
Second order  & $\mU \mSigma^{\tfrac{1}{4}}\mV^{\top}$   & 1e-2 & 0 & 2.877 \\
Second order  & $\mU \mSigma^{\tfrac{1}{4}}\mV^{\top}$   & 2e-2 & 0 & \textbf{2.872} \\
Second order  & $\mU \mSigma^{\tfrac{1}{4}}\mV^{\top}$   & 3e-2 & 0 & 2.873 \\
Second order  & $\mU \mSigma^{\tfrac{1}{4}}\mV^{\top}$   & 4e-2 & 0 & 2.876 \\
\hline 

Second order  & $\mU \mSigma^{0}\mV^{\top}$   & 1e-3 & 0 & 2.921 \\
Second order  & $\mU \mSigma^{0}\mV^{\top}$   & 3e-3 & 0 & 2.887 \\
Second order  & $\mU \mSigma^{0}\mV^{\top}$   & 6e-3 & 0 & 2.880 \\
Second order  & $\mU \mSigma^{0}\mV^{\top}$   & 7e-3 & 0 & \textbf{2.879} \\
Second order  & $\mU \mSigma^{0}\mV^{\top}$   & 8e-3 & 0 & 2.880 \\
Second order  & $\mU \mSigma^{0}\mV^{\top}$   & 1e-2 & 0 & 2.880 \\
\hline 
\bottomrule
\end{tabular}
\end{table}

\end{document}